\documentclass[journal]{IEEEtran}
\usepackage{ulem}
\usepackage{float}
\usepackage{algorithm}
\usepackage{algpseudocode}
\usepackage[marginal]{footmisc}
\usepackage[numbers,square,comma,sort&compress]{natbib}

\usepackage{graphics} 
\usepackage{epsfig} 
\usepackage{mathptmx} 
\usepackage{times} 
\usepackage{indentfirst}
\usepackage{amsmath}
\usepackage{bbding}
\usepackage{amssymb}  
\usepackage{booktabs}
\usepackage{url}
\usepackage{multirow}
\usepackage{bbding}
\usepackage{hyperref}
\usepackage{xcolor}
\usepackage{blindtext}

\usepackage{etoolbox}
\makeatletter
\patchcmd{\@makecaption}
  {\scshape}
  {}
  {}
  {}
\makeatletter
\patchcmd{\@makecaption}
  {\\}
  {.\ }
  {}
  {}
\makeatother

\ifCLASSINFOpdf
\else
\fi
\begin{document}
%
\title{Enhance 3D Visual Grounding through LiDAR and Radar Point Clouds Fusion for Autonomous Driving}
%
%
%

\author{Runwei Guan,
        Jianan Liu,
        Ningwei Ouyang,
        Shaofeng Liang,
        Daizong Liu, 
        Xiaolou Sun,
        Lianqing Zheng,\\
        Ming Xu,
        Tao Huang,
        Yutao Yue$^\dagger$,
        Guoqiang Mao,~\IEEEmembership{Fellow,~IEEE},
        and Hui Xiong$^\dagger$,~\IEEEmembership{Fellow,~IEEE}
\thanks{Runwei Guan and Jianan Liu are co-first authors.}
\thanks{Corresponding author: Yutao Yue and Hui Xiong.}
\thanks{Runwei Guan, Yutao Yue and Hui Xiong are with Thrust of Artificial Intelligence, Hong Kong University of Science and Technology (Guangzhou), China. (\{runwayrwguan, yutaoyue, xionghui\}@hkust-gz.edu.cn)}
\thanks{Jianan Liu is with Mononai AI, Sweden. (jianan.liu@momoniai.org)}
\thanks{Shaofeng Liang is with College of Computer Science and Technology, China University of Petroleum (East China), China. (S23070053@s.upc.edu.cn)}
\thanks{Ningwei Ouyang and Ming Xu are with School of Advanced Technology, Xi'an Jiaotong-Liverpool University, China. (ningwei.ouyang@liverpool.ac.uk, ming.xu@xjtlu.edu.cn)}
\thanks{Daizong Liu is with Wangxuan Institute of Computer Technology, Peking University, China. ( dzliu@stu.pku.edu.cn)}
\thanks{Xiaolou Sun is with School of Automation, Southeast University, China. (xlsun@seu.edu.cn)}
\thanks{Lianqing Zheng is with School of Automotive Studies, Tongji University (zhenglianqing@tongji.edu.cn)}
\thanks{Tao Huang is with College of Science and Engineering, James Cook University, Australia. (tao.huang1@jcu.edu.au)}
\thanks{Guoqiang Mao is with School of Transportation, Southeast University, China. (g.mao@ieee.org)}
}

%
%

\markboth{submitted to IEEE Transactions on Intelligent Transportation Systems}%
{Shell \MakeLowercase{\textit{et al.}}: Bare Demo of IEEEtran.cls for IEEE Journals}
%



\maketitle

\begin{abstract}
Embodied outdoor scene understanding forms the foundation for autonomous agents to perceive, analyze, and react to dynamic driving environments. However, existing 3D understanding is predominantly based on 2D Vision-Language Models (VLMs), which collect and process limited scene-aware contexts. In contrast, compared to the 2D planar visual information, point cloud sensors such as LiDAR provide rich depth and fine-grained 3D representations of objects. Even better the emerging 4D millimeter-wave radar detects the motion trend, velocity, and reflection intensity of each object. 
The integration of these two modalities provides more flexible querying conditions for natural language, thereby supporting more accurate 3D visual grounding. To this end, we propose a novel method called TPCNet, the first outdoor 3D visual grounding model upon the paradigm of prompt-guided point cloud sensor combination, including both LiDAR and radar sensors. To optimally combine the features of these two sensors required by the prompt, we design a multi-fusion paradigm called Two-Stage Heterogeneous Modal Adaptive Fusion. Specifically, this paradigm initially employs Bidirectional Agent Cross-Attention (BACA), which feeds both-sensor features, characterized by global receptive fields, to the text features for querying. Moreover, we design a Dynamic Gated Graph Fusion (DGGF) module to locate the regions of interest identified by the queries. To further enhance accuracy, we devise an C3D-RECHead, based on the nearest object edge to the ego-vehicle. Experimental results demonstrate that our TPCNet, along with its individual modules, achieves the state-of-the-art performance on both the Talk2Radar and Talk2Car datasets. We release the code at \url{https://github.com/GuanRunwei/TPCNet}.
\end{abstract}
\vspace{-4mm}

%
\IEEEpeerreviewmaketitle

\section{Introduction}
With the rapid advancement of Autonomous Driving (AD), 3D perception based on multi-sensor fusion has been widely adopted in Autonomous Vehicles (AVs), robotics, and roadside perception systems \cite{song2024robustness}. As a primary sensor for 3D perception, LiDAR provides accurate positioning \cite{meng2024traffic} and detailed 3D representations of objects \cite{chen2025leveraging}, and has been demonstrated to be suitable for numerous perception tasks in AD \cite{zhang2023fs}. However, LiDAR cannot capture crucial information such as the motion trend and velocity \cite{liu2025mssf}, which are essential for understanding surrounding entities \cite{choi2021road}. In contrast, the mmWave radar (radar) is capable of sensing the distance and orientation of each object, with a detection range that exceeds that of LiDAR \cite{engels2021automotive}. Moreover, radar can capture motion and velocity while operating reliably under adverse weather. The latest 4D radar offers richer point cloud data when compared with conventional 2D radar, showing significant potential for 3D perception as well \cite{hasan2024mm}. Consequently, considerable research has focused on the fusion of LiDAR and radar for enhanced 3D perception \cite{yang2024ralibev}. 

\begin{figure}
    \centering
    \includegraphics[width=0.998\linewidth]{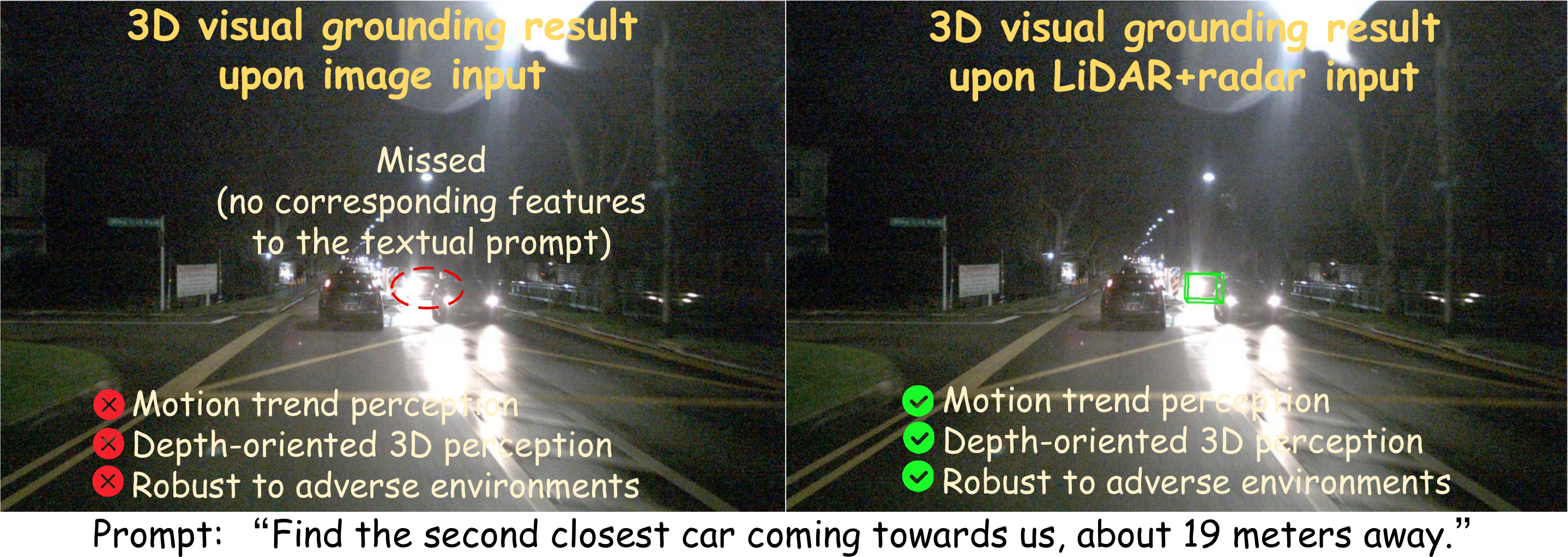}
    \vspace{-5mm}
    \caption{Comparison between camera-based 2D visual grounding and LiDAR-radar-based 3D visual grounding. Under adverse conditions, objects referred to by text prompts containing depth (distance) and motion cues can be localized within the contextual scene using point clouds (LiDAR + radar), which cannot be achieved by cameras.}
    \label{fig:first_compare}
\end{figure}

\begin{figure*}
    \centering
    \includegraphics[width=0.92\linewidth]{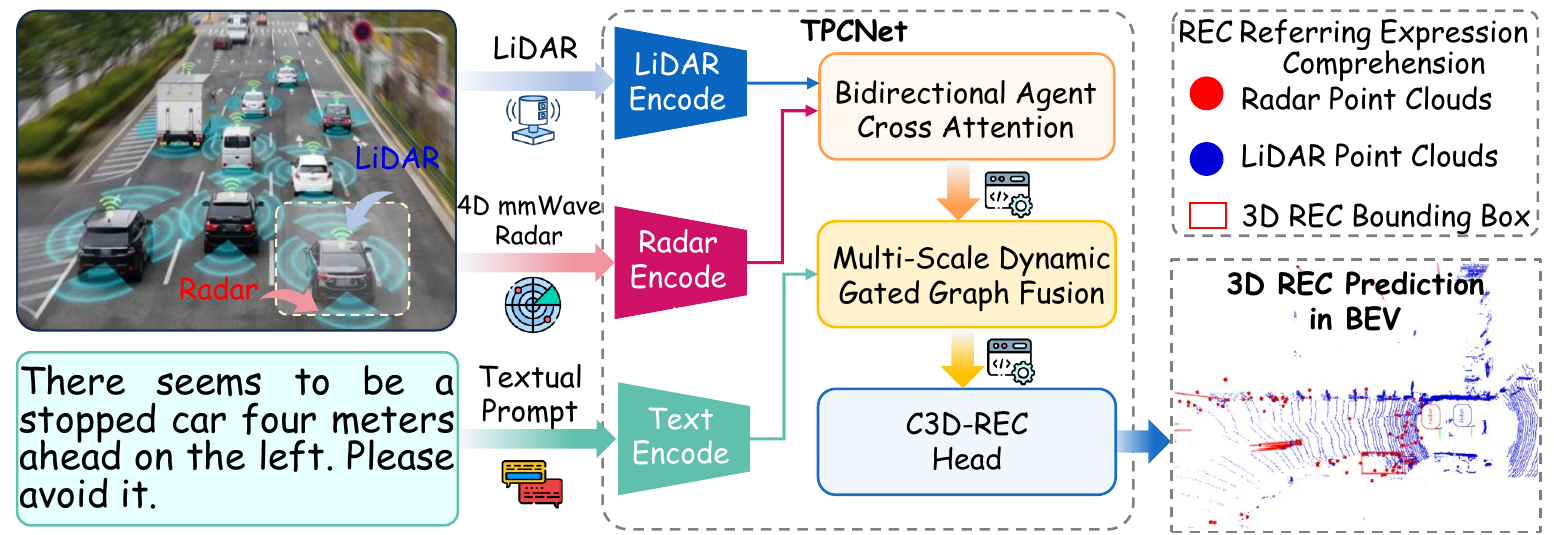}
    \vspace{-3mm}
    \caption{Overview of Talk2PC pipeline, where the textual prompt guides LiDAR and 4D mmWave radar (radar) to localize the referred object(s).}
    \label{fig:overview}
\end{figure*}

Besides, as Vision-Language Models (VLMs) are progressively applied to interactive perception and embodied intelligence, they enable AVs and robots not only to perceive scenes but also to understand human intentions and locate corresponding objects \cite{zhang2025clip}. Nonetheless, the current advancements have predominantly focused on the integration with visual modalities, while limited attention has been caught with 3D point cloud sensors, particularly in the domain of language-guided multi-sensor fusion for 3D visual grounding in traffic scenarios. For instance, Talk2Car \cite{deruyttere2019talk2car} proposes a benchmark of 2D visual grounding on the image plane from the perspective of a driving car. WaterVG \cite{guan2024watervg} focuses on the 2D visual grounding based on camera-radar fusion.  Cheng~et al. \cite{cheng2023language} extended Talk2Car to LiDAR-based 3D visual grounding while Talk2Radar \cite{guan2024talk2radar} explores the capacity of 4D radars on 3D visual grounding. It is therefore evident that natural language querying of objects offers a more flexible and intuitive interactive paradigm for open-world traffic perception and intelligent transportation systems. However, as illustrated in Fig. \ref{fig:first_compare}, we observe that with the increasing demands of environmental perception, the 2D image plane is no longer sufficient for queries involving specific object attributes such as depth, motion trends, or perception under severely adverse conditions. For instance, when querying ``a vehicle approximately 20 meters ahead, moving toward the ego-vehicle at a speed of 30 km/h”, an image alone fails to capture any of the object characteristics required by such a natural language prompt. Addressing these queries necessitates accurate 3D scene geometry modeling and a deeper understanding of the physical properties of traffic participants.


To address the aforementioned challenges, an exploratory approach is introduced in this paper that leverages language guidance for LiDAR and radar fusion within a dual-sensor framework for 3D visual grounding, as briefly illustrated in Fig. \ref{fig:overview}. Given the inherent variability in object attribute descriptions within prompts, a dynamic weighting mechanism is required to adaptively adjust the relative contributions of LiDAR and radar information based on the given prompt.

Unlike existing methods, including Adaptive Fuson Module \cite{xu2024rlnet}, Query-based Interactive Module \cite{yang2024ralibev} and InternRAL \cite{wang2023bi}  that employ static fusion strategies using convolution-based approaches for LiDAR and radar integration, we propose a novel Bi-directional Agent Cross Attention (BACA) mechanism. This approach enables multi-scale feature fusion in the encoder stage, alternating between LiDAR and radar as the primary information source for query processing. Notably, BACA substantially reduces computational complexity compared to conventional cross-attention mechanisms while enabling efficient dynamic modeling.

Furthermore, to mitigate false positives arising from point cloud sensors in 3D visual grounding and to effectively correlate multiple point cloud objects referenced within a prompt, we introduce a Dynamic Gated Graph Fusion (DGGF) strategy. A key limitation of the previous Gated Graph Fusion (GGF) approach \cite{guan2024talk2radar} lies in its reliance on a static graph construction method, where all features contribute to the same graph without considering node similarity, thereby preventing adaptive graph structure adjustments based on image content. This constraint diminishes the advantages of graph-based models. In contrast, DGGF employs a linguistically conditioned candidate region modeling process, utilizing a dynamic axial graph neural network with feature gating to enhance context-awareness and adaptability.

Additionally, to address prediction errors caused by the absence of object center point clouds and to ensure depth information alignment with prompt descriptions, we propose a 3D visual grounding prediction head, termed Corner3D-RECHead (C3D-RECHead). This head identifies the region with the denser point cloud near the corner of the object closest to the ego vehicle, utilizing this as the anchor centroid to enhance localization accuracy.

The main contributions are summarized as below:
\begin{itemize}
\item We propose TPCNet, which performs 3D visual grounding by leveraging textual semantics to guide dual point cloud sensors. TPCNet adaptively and dynamically aligns and fuses linguistic features with heterogeneous point clouds. It achieves state-of-the-art performance on both the Talk2Radar and Talk2Car datasets.
\item We propose the Bi-directional Agent Cross Attention (BACA) to fuse the features of LiDAR and radar with sufficient consideration of respective physical characteristics, which enables TPCNet to embed radar representations of object motion alongside LiDAR's 3D geometric understanding of the environment.
\item We propose Dynamic Gated Graph Fusion (DGGF), which constructs a dynamic axial graph for the adaptive fusion of point cloud features in the linguistic feature space, while filtering regions of interest.
\item In alignment with the physical properties of point cloud sensors, we propose a 3D visual grounding head called C3D-RECHead, centered on the closest edge between the object and the ego-vehicle. C3D-RECHead effectively reduces the object miss detection rate while significantly enhancing localization precision in samples where the prompt includes numeric depth information.
\end{itemize}

The remained content is organized as follows: Section \ref{sec:related} states the related works; Section \ref{sec:method} illustrates proposed methods; Section \ref{sec:experiments} demonstrates the experiments; Section \ref{sec:conclusion} concludes the paper and lists the limitation and future works.

\begin{figure*}
    \includegraphics[width=0.99\linewidth]{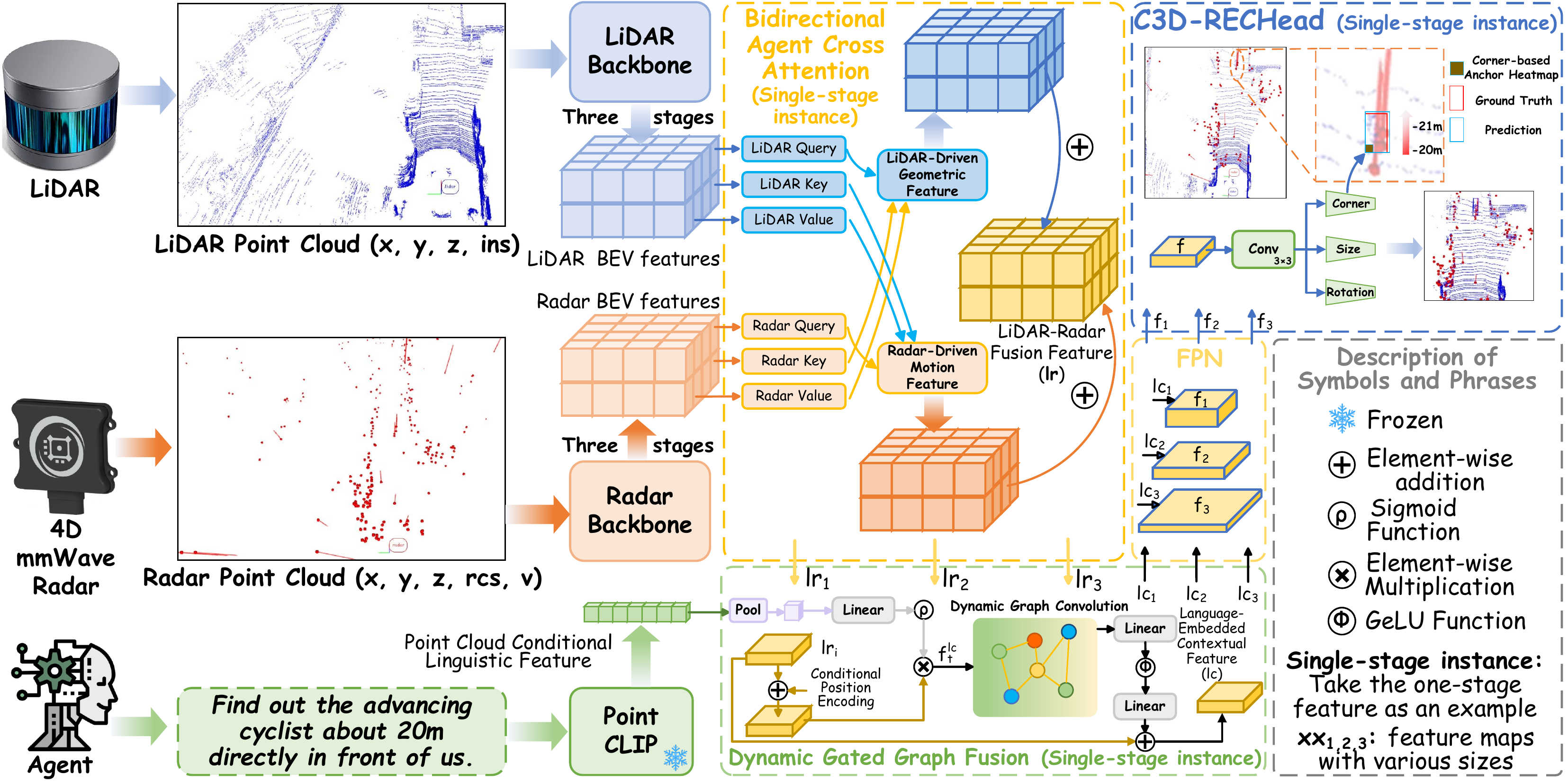}
    \vspace{-3mm}
    \caption{The architecture of proposed TPCNet. In TPCNet, Bidirectional Agent Cross Attention (BACA), Dynamic Gated Graph Fusion (DGGF) and C3D-RECHead are three core modules, which are illustrated in detail in Sub-section \ref{subsec:baca}, \ref{subsec:dggf} and \ref{subsec:c3d}.}
    \label{fig:model}
\end{figure*}

\section{Related Works}
\label{sec:related}

\subsection{3D Object Detection with LiDAR-Radar Fusion}
Leveraging the high-precision 3D sensing capabilities of LiDAR and the all-weather robustness and long-range sensing advantages of radar, the fusion of these two modalities has been widely recognized as a complementary approach for 3D object detection. Several studies have introduced bi-directional radar-LiDAR fusion modules, utilizing convolution-based attention mechanisms that compute the product of feature maps to dynamically weight the respective modality features. Among them, Wang et al. \cite{wang2022multi} propose InterRAL, which fuses the features of LiDAR and radar through the softMax gating mechanism. Yang et al. \cite{yang2024ralibev} introduce a query-based interactive feature fusion module to concatenate feature maps of selected LiDAR and radar points. Huang et al. \cite{huang2024l4dr} propose Multi-Scale Gated Fusion module to counteract the varying degrees of sensor degradation. Xu et al. \cite{xu2024rlnet} propose adaptive fusion module to select salient features of LiDAR and radar. Further advancements in this domain include the cross-fusion approach introduced by Meng et al. \cite{meng2024traffic}, which generates pseudo-radar features guided by LiDAR information. Similarly, Wang et al. \cite{wang2023bi} developed a two-stage LiDAR-radar fusion framework to extract high-quality composite point cloud features.

Building upon these developments, we identify two key trends: (i) several methods adopt LiDAR-guided radar fusion to explore the complementary features of both modalities and (ii) some approaches utilize bi-directional fusion strategies to compensate for missing modality-specific information while simultaneously refining object representations. However, these existing methods predominantly rely on static convolution-based techniques, which exhibit limited generalization capability when applied across varying sensor configurations and environmental conditions.

To address this limitation, recent studies have explored attention-based fusion mechanisms to enable dynamic cross-attention and adaptive information exchange between modalities. Qian et al. \cite{qian2021robust} propose a dual-branch attention module to dynamically weigh the significance of LiDAR and radar features. Chae et al. \cite{chae2024towards} introduce the LRF module, which performs LiDAR-based query to match the radar feature by cross attention. Nevertheless, these attention-driven approaches often operate on high-dimensional feature spaces, causing a quadratic increase in computational complexity and a significant model parameter overhead, which compromises their representational efficiency.

To overcome these challenges, we introduce the Bi-directional Agent Cross Attention (BACA) mechanism, which achieves linear computational complexity while leveraging LiDAR and radar as dual sources of 3D geometric context and motion features. This approach ensures the extraction of rich, dynamically aligned features that effectively correspond to textual prompts, enhancing the adaptability and efficiency of multi-modal 3D object detection.

\subsection{3D Outdoor Visual Grounding}
Visual grounding, a vision-language task aiming to localize objects based on natural language prompts \cite{ke2025graph}, has gained significant attention in outdoor applications ranging from autonomous vehicles to embodied intelligence \cite{qian20223d}. Recent advances extend beyond traditional camera-based 2D grounding to multi-sensor and 3D spatial understanding \cite{zhang2025clip}. 

In 2D domain, \cite{guan2024watervg} established a multi-task benchmark with two-stage fusion of camera, radar, and language features. However, these approaches remain limited to 2D planes and lack comprehensive 3D spatial reasoning capabilities. The field has recently evolved toward 3D outdoor grounding, driven by autonomous navigation demands. Cheng~et al. \cite{cheng2023language} pioneered LiDAR-based 3D grounding for driving scenarios, whereas Zhan~et al. \cite{zhan2024mono3dvg} proposed a monocular-based 3D REC baseline model called Mono3DVG-TR. Radar-based solutions have also emerged, with Guan~et al. \cite{guan2024talk2radar} establishing a 4D radar benchmark. 

Despite these advancements, the current research exhibits three limitations: (1) insufficient integration of complementary sensors, (2) limited adaptability to dynamic textual guidance, and (3) under-explored human-robot interaction paradigms for vehicles. Our work bridges these gaps through two key contributions. First, we introduce a unified 3D grounding framework to support point cloud sensors (e.g., LiDAR and radar) with modality-adaptive fusion. Second, we develop dynamic attention mechanisms that automatically adjust sensor weighting based on textual semantics, outperforming conventional sequential fusion approaches. The proposed Talk2PC system demonstrates superior performance in both autonomous driving scenarios and interactive robot navigation tasks.

\section{Methodology}
\label{sec:method}

\subsection{The Overall Pipeline}
\label{subsec:pipeline}
Fig. \ref{fig:model} illustrates the overall architecture of the proposed TPCNet. TPCNet processes inputs from two complementary perception modalities: LiDAR and 4D mmWave radar, both represented as point clouds. Each LiDAR point comprises 3D spatial coordinates \((x, y, z)\) and intensity \((ins)\), while each radar point includes 3D coordinates (aligned to the LiDAR frame), radar cross-section \((rcs)\), and compensated radial velocity \((v)\). In addition, TPCNet receives textual prompts from an agent to guide object localization within the scene.

LiDAR and radar point clouds are independently encoded by a pillar-based backbone, producing multi-scale features at three resolutions: LiDAR features \(f^l_{i}\) and radar features \(f^r_{i}\), where \(i \in \{1,2,3\}\). Textual instructions are encoded via a language encoder to obtain semantic embeddings.

Multi-scale LiDAR and radar features are fused by the proposed Bidirectional Agent Cross Attention (BACA) module, yielding LiDAR–radar fusion features \(lr_{i}\). These are further integrated with textual embeddings through the Dynamic Gated Graph Fusion (DGGF) module to produce language-conditioned contextual features \(lc_{i}\). 

The resulting features are refined via a Feature Pyramid Network (FPN) to enhance multi-scale representation, producing \(f_{i}\). Finally, the proposed C3D-RECHead predicts the queried object’s 3D position, size, and orientation from the fused features, conditioned on the textual prompt.

\subsection{Backbones of LiDAR, Radar and Textual Instruction}
\label{subsec:backbone}
During feature encoding, LiDAR and radar point clouds are independently processed by pillar-based backbones~\cite{lang2019pointpillars}, generating Bird’s-Eye-View (BEV) pillar features \( f^l \) and \( f^r \). Each backbone produces three-stage feature maps, with each stage represented as \(\{f^l, f^r\} \in \mathbb{R}^{C \times H \times W}\). For textual instructions, we adopt the text encoder of PointCLIP~\cite{zhang2022pointclip} rather than standalone language models (e.g., BERT~\cite{devlin2019bert}), as the latter lack point cloud awareness. PointCLIP exploits the contrastive learning framework of CLIP~\cite{radford2021learning} to embed both point clouds and text into a unified representation space, facilitating robust cross-modal alignment. This yields textual features \( pl \in \mathbb{R}^{C \times L} \).



\begin{figure}
\centering
    \includegraphics[width=0.95\linewidth]{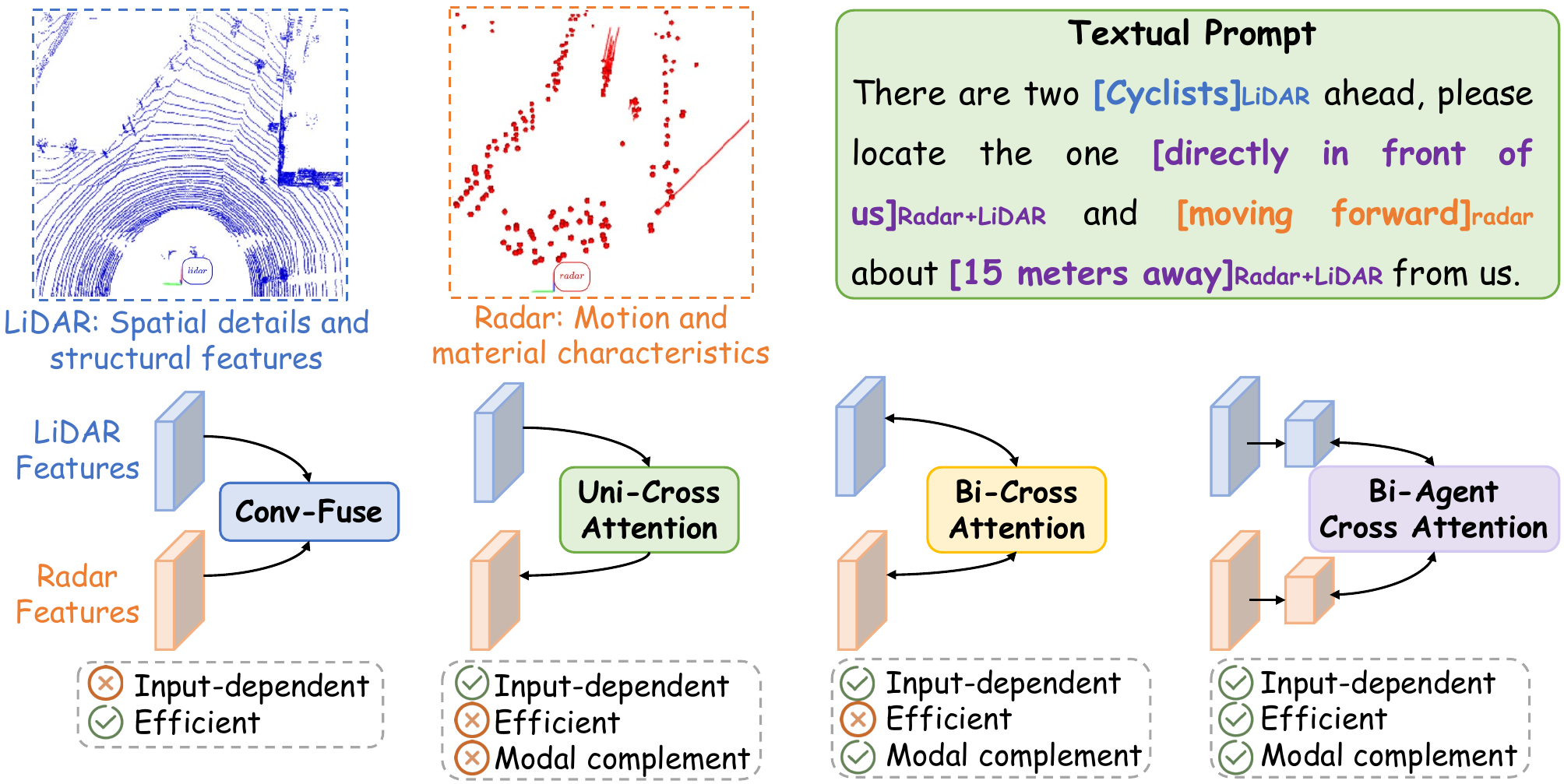}
    \vspace{-3mm}
    \caption{The significance of BACA for 3D visual grounding based on the fusion of LiDAR and radar.}
    \label{fig:baca_simple}
\end{figure}

\subsection{Bidirectional Agent Cross Attention}
\label{subsec:baca}

\begin{figure}
\centering
    \includegraphics[width=0.94\linewidth]{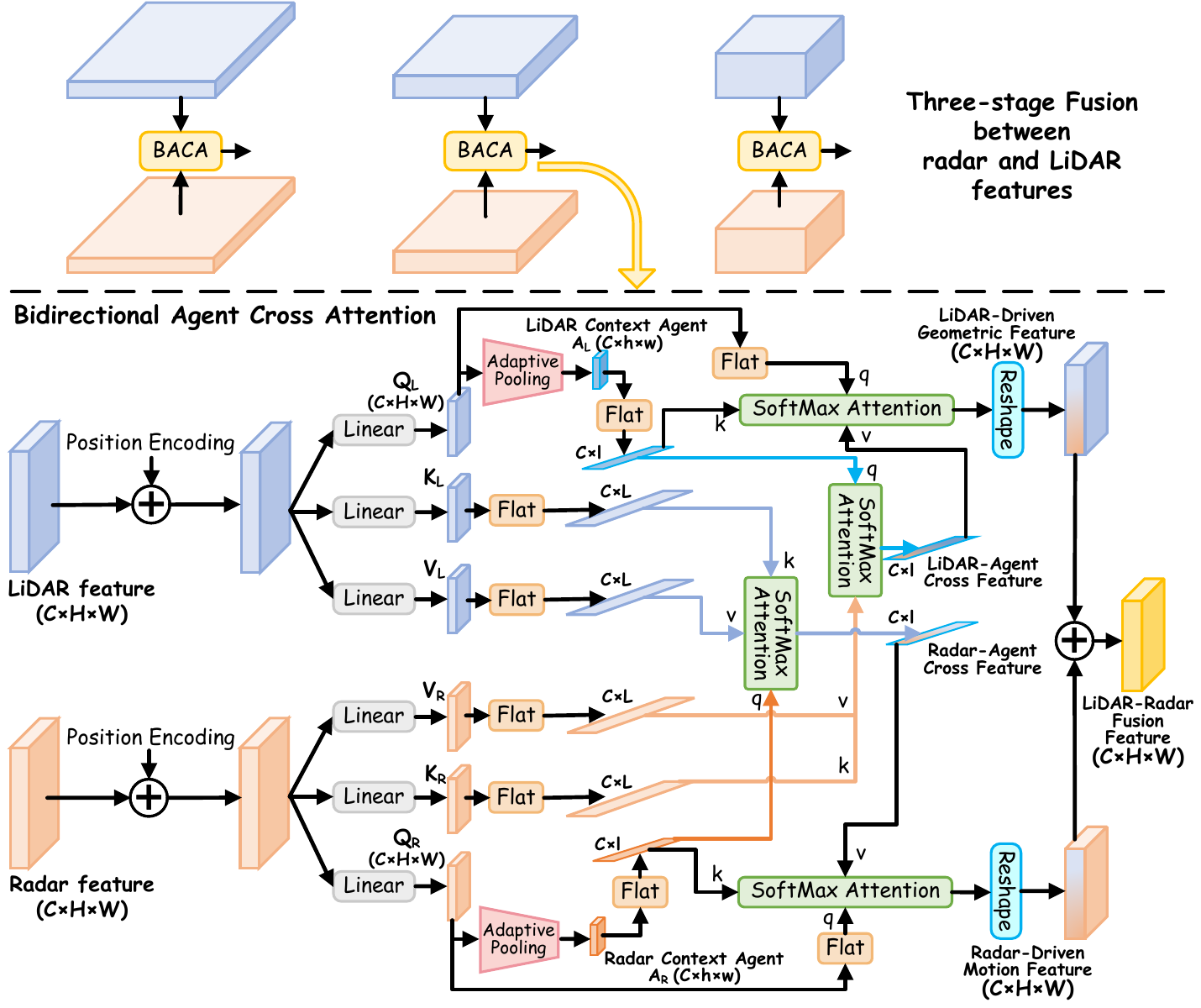}
    \vspace{-4mm}
    \caption{The detailed structure of Bidirectional Agent Cross Attention for the fusion of LiDAR and radar feature. Here is the instance for one-stage features.}
    \label{fig:baca}
\end{figure}

As shown in Fig. \ref{fig:baca_simple}, LiDAR offers high-precision 3D spatial measurements, whereas radar is robust to adverse weather and excels at capturing object motion. Fusing these complementary modalities enables a more complete scene understanding. However, textual descriptions may omit certain attributes, such as semantics, velocity, direction, or depth, making it essential for dual sensors to dynamically extract and align features with the textual instructions. In dual-sensor settings, fully exploiting the context from each modality and enabling reciprocal feature exchange mitigates the asymmetry inherent in unidirectional fusion, while allowing mutual verification between sensors. Motivated by these factors, we introduce the Bidirectional Agent Cross Attention (BACA) module, an efficient mechanism for rapid, bidirectional LiDAR–radar feature fusion conditioned on textual guidance.

As depicted in Fig. \ref{fig:baca}, the features from LiDAR \(f^l \in \mathbb{R}^{C \times H \times W}\) and Radar \(f^r \in \mathbb{R}^{C \times H \times W}\) are initially augmented with positional encoding \(\mathtt{PE}\) \cite{vaswani2017attention}. These enhanced features are subsequently passed through three linear feedforward modules to generate triplet features. In this context, \(Q_L\), \(K_L\), and \(V_L\) represent the query, key, and value matrices for the LiDAR, respectively, while \(Q_R\), \(K_R\), and \(V_R\) denote the corresponding matrices for the radar. The initialization process described above is detailed in Equations (\ref{eq:pos_encoding_1}) to (\ref{eq:pos_encoding_6}) as below:

\begin{align}
    & Q_L = (f^l+\mathtt{PE}) \mathbf{W}_{QL},  Q_L \in \mathbb{R}^{C \times H \times W},
    \label{eq:pos_encoding_1} \\
    & K_L = \mathtt{Flat}((f^l+\mathtt{PE}) \mathbf{W}_{KL}), K_L \in \mathbb{R}^{C \times L \ (L=H \times W)}, 
    \label{eq:pos_encoding_2} \\
    & V_L = \mathtt{Flat}((f^l+\mathtt{PE}) \mathbf{W}_{VL}),  V_L \in \mathbb{R}^{C \times L \ (L=H \times W)},
    \label{eq:pos_encoding_3}
\end{align}

\begin{align}
    & Q_R = (f^r +\mathtt{PE}) \mathbf{W}_{QR},  Q_R \in \mathbb{R}^{C \times H \times W},
    \label{eq:pos_encoding_4} \\
    & K_R = \mathtt{Flat}((f^r +\mathtt{PE}) \mathbf{W}_{KR}), K_R \in \mathbb{R}^{C \times  L \ (L=H \times W)}, 
    \label{eq:pos_encoding_5} \\
    & V_R = \mathtt{Flat}((f^r +\mathtt{PE}) \mathbf{W}_{VR}),  V_R \in \mathbb{R}^{C \times  L \ (L=H \times W)},
    \label{eq:pos_encoding_6}
\end{align}
where $C$ is the number of channel while $H$ and $W$ denotes the height and width of the feature map. $L$ is the product result of $H$ and $W$. $\mathtt{Flat}(\cdot)$ denotes the flatting operation along the spatial dimension for the image-like feature map.

Furthermore, we generate the context agent features of LiDAR and radar, $A_L \in \mathbb{R}^{C \times h \times w}$ and $A_R \in \mathbb{R}^{C \times h \times w}$, by exerting the adaptive pooling to the two query matrices $Q_L$ and $Q_R$, where two context agent features $A_L$ and $A_R$ maintain the basic contextual structures while reducing the computational cost for the downstream calculation on the softmax attention. Based on the above, we leverage softmax attention $\mathtt{Attn}(\cdot)$ to fuse the features of two sensors. Exactly, for the generation process of LiDAR-driven geometric feature $f_{lg}$, the lidar context agent feature $A_L$ firstly serves as the agent of $Q_L$, which aggregates the global contextual information in $K_R$ and $V_R$ provided by radar. Based on the above, we obtain the LiDAR-Agent Cross Feature $f_{lc}$. Subsequently, we take $A_L$ as the key while $f_{lc}$ as the value in the second softmax attention with the orginal $Q_L$ as the query. Here, we broadcast the global contextual information upon the agent feature $A_L$ to every query tokens, where such process avoids the direct calculation of pairwise similarities between query and key features, maintaining the information exchange via agent feature. Likewise, the generation process of Radar-driven motion feature $f_{rm}$ follows the same process with LiDAR-driven geometric feature $f_{lg}$. The whole process is shown in Eq. (\ref{eq:baca_1}) and (\ref{eq:baca_2}).

\begin{align}
\left\{
    \begin{aligned}
    & A_L = \mathtt{Flat}(\mathtt{AdaPool}(Q_L)), A_L \in \mathbb{R}^{C \times l}, \\
    & f_{lc} = \mathtt{Attn}(A_L, K_R, V_R) = \frac{A_L K_R^T}{\sqrt{d}}\cdot V_R, f_{lc} \in \mathbb{R}^{C \times l}, \\
    & f_{lg} = \mathtt{Attn}(Q_L, A_L, f_{lc}) = \frac{Q_L A_L^T}{\sqrt{d}}\cdot f_{lc}, f_{lg} \in \mathbb{R}^{C \times L}, \\
    & f_{lg} = \mathtt{Reshape}(f_{lg}), f_{lg} \in \mathbb{R}^{C \times H \times W},
    \label{eq:baca_1} \\
    \end{aligned}
    \right.
    \\
    \left\{
    \begin{aligned}
        & A_R = \mathtt{Flat}(\mathtt{AdaPool}(Q_R)), A_R \in \mathbb{R}^{C \times l}, \\
        & f_{rc} = \mathtt{Attn}(A_R, K_L, V_L) = \frac{A_R K_L^T}{\sqrt{d}}\cdot V_L, f_{rc} \in \mathbb{R}^{C \times l}, \\
        & f_{rm} = \mathtt{Attn}(Q_R, A_R, f_{rc}) = \frac{Q_R A_R^T}{\sqrt{d}}\cdot f_{rc}, f_{rm} \in \mathbb{R}^{C \times L}, \\
        & f_{rm} = \mathtt{Reshape}(f_{rm}), f_{rm} \in \mathbb{R}^{C \times H \times W},
        \label{eq:baca_2} 
    \end{aligned}
\right.
\end{align}
where $\mathtt{AdaPool}(\cdot)$ represents the adaptive pooling operation. $l$ is the product result of $h$ and $w$, where $h$ and $w$ denote the height and width of the agent feature. The values of $h$ and $w$ are both smaller than $H$ and $W$. In this context, the size of the agent feature $l$ is a hyperparameter, significantly smaller than the original LiDAR and Radar feature dimensions. It achieves a linear computational complexity of \( O(LlC) \). The complexity of our proposed BACA is much smaller than vanilla cross attention's \( O(L^2 C) \) while still preserving the global cross-modal fusion capability.

\subsection{Dynamic Gated Graph Fusion}
\label{subsec:dggf}

To efficiently fuse the features of point clouds and textual prompts, as shown in Fig. \ref{fig:model}, we employ a cross-modal gating mechanism to integrate the language and point cloud features, and construct dynamic graph aggregation to capture the regions of interest within the point cloud context. 

Let the linguistic feature provided by PointCLIP be denoted as \( f_t \). We first apply the Max-Pooling operation to obtain the compressed linguistic feature \( f_t^s \), which then passes through a feedforward layer with a sigmoid activation function to compute the gating weight \( \mathbf{W}_G \). In parallel, the LiDAR-Radar fusion feature \( lr \) is first augmented with Conditional Position Encoding (\( \mathtt{CPE} \)), and the resulting position-aware feature is element-wise multiplied by the gating weight \( \mathbf{W}_G \), yielding the language-conditioned point cloud feature \( f_t^{lc} \). The detailed process is shown in below:
\begin{align}
    & f_t^s = \mathtt{MaxPool}(f_t), \label{eq:text_point_fusion1} \\
    & \mathbf{W}_G = \rho(\mathbf{W}\cdot f_t^s), \label{eq:text_point_fusion2} \\
    & f_t^{lc} = \mathbf{W}_G \cdot (lr + \mathtt{CPE}(lr)), \label{eq:text_point_fusion3}
\end{align}
where $\rho(\cdot)$ denotes the Sigmoid function.

Secondly, to effectively capture regions of interest (RoIs) within the point cloud based on the textual prompt, while preserving surrounding contextual information, we employ graph-based modeling, which is well-suited for handling complex spatial and structural relationships. Since point cloud data lacks a fixed grid structure, we represent features as a set of nodes within a graph, enhancing the capability to model both spatial distributions and semantic attributes of objects. 

\begin{figure}
\centering
    \includegraphics[width=0.99\linewidth]{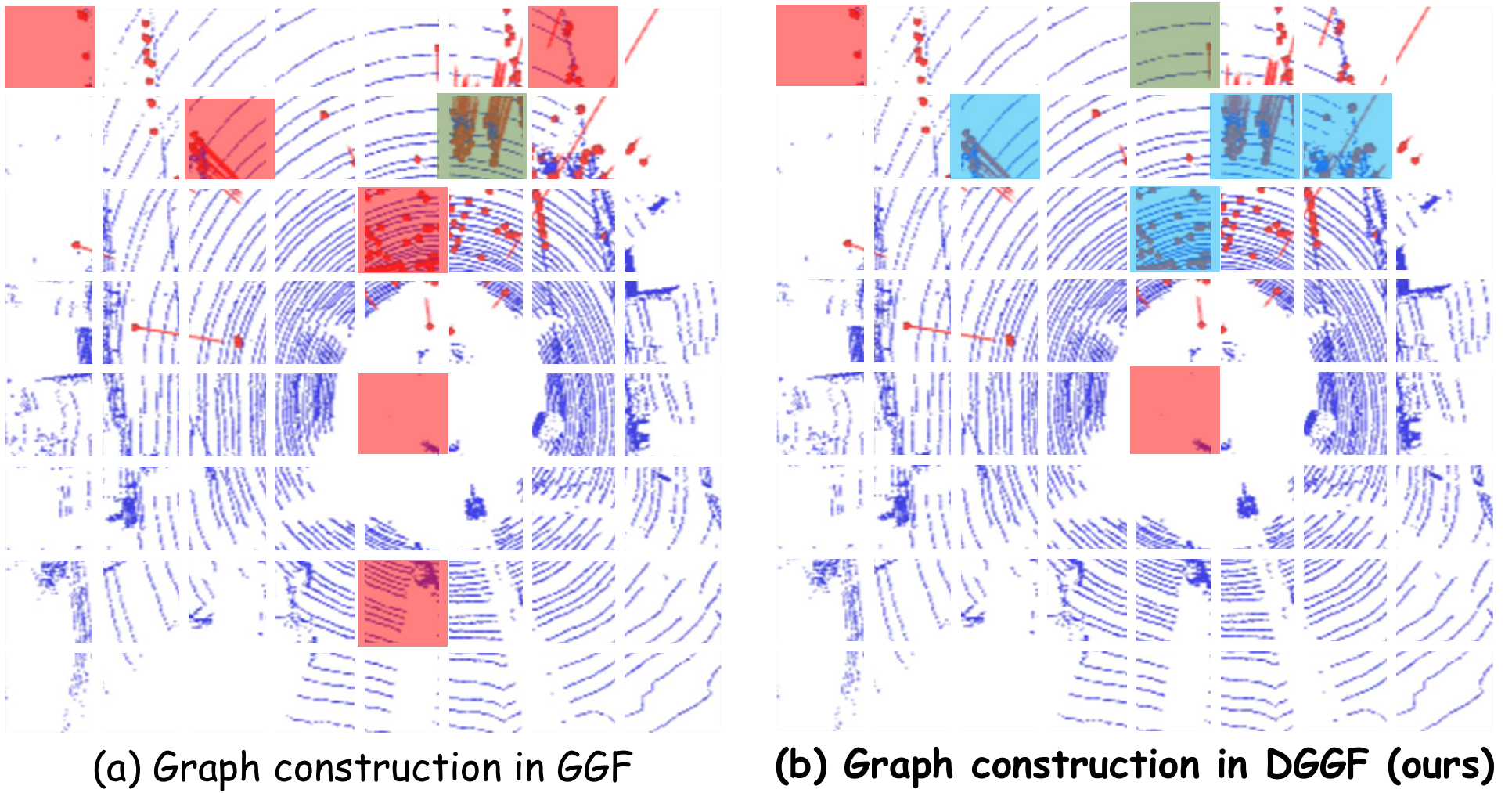}
    \vspace{-6mm}
    \caption{The construction of the dynamic graph in DGGF. (a) The graph construction in GGF for the green patch of an 8 $\times$ 8 feature map. All red regions will be linked to the green region, irrespective of their similarity. (b) Dynamic graph construction applied to the green region of an 8 $\times$ 8 feature map, which adaptively builds a graph along the axes by utilizing a mask (represented by the blue regions), and ensures that only patches with similar Euclidean distances are connected. The red patches remain disconnected from the green patch, as they do not fall within the scope of the mask.}
    \label{fig:dynamic_graph_construction}
\end{figure}

As illustrated in Fig. \ref{fig:dynamic_graph_construction}, compared to the static graph constructed in Sparse Vision Graph Attention (SVGA) \cite{munir2023mobilevig}, which remains unchanged across different features, we propose a dynamic graph that retains SVGA’s axial structure efficiency while dynamically adjusting graph connectivity based on feature characteristics. To achieve this, rather than computing all pairwise distances exhaustively, we estimate the mean ($\mu$) and standard deviation ($\sigma$) of Euclidean distances between nodes by analyzing a subset of nodes, which is derived by partitioning the image into quadrants and evaluating the pairwise correspondence between diagonally opposite regions. Specifically, we partition the pseudo-image feature map of the point cloud into quadrants and compare diagonal pairs within these quadrants. This process provides an efficient approximation of $\mu$ and $\sigma$, significantly reducing the computational cost associated with exhaustive distance calculations.  

Direct computation of exact $\mu$ and $\sigma$ values would require calculating Euclidean distances between every node pair, causing excessive computational complexity. Instead, we approximate these values using the aforementioned subset analysis, which balances efficiency and accuracy. Subsequently, we adopt row and column-wise connections, following SVGA, to further reduce computational overhead. As demonstrated by MobileViG \cite{munir2023mobilevig}, not all nodes need to be connected; instead, if the Euclidean distance between nodes is less than the estimated difference between $\mu$ and $\sigma$, a connection is established.  

Unlike conventional GNN approaches, where a fixed number of K-nearest neighbors (KNN) is used for all images, our DGGF approach supports a variable number of connections across different point cloud feature maps. This flexibility allows adaptive connectivity, ensuring that node relationships better reflect the underlying feature distributions. The rationale for leveraging $\mu$ and $\sigma$ is that nodes within the range of $\mu \pm \sigma$ are likely to be spatially correlated and should exchange information. These values are subsequently used to establish graph connections, as illustrated in Fig. \ref{fig:dynamic_graph_construction} (b).  

As shown in Fig. \ref{fig:dynamic_graph_construction}, the GGF tends to connect RoIs with non-object regions, potentially introducing irrelevant information. In contrast, our proposed DGGF selectively connects object regions within the point cloud feature map, ensuring a more precise and contextually relevant feature representation.

\begin{algorithm}
\footnotesize
\caption{Algorithm of Dynamic Gragh Convolution ($\mathtt{DynConv}$) in DGGF}\label{alg:dynamic_graph}
\begin{algorithmic}
\State \textbf{Input:} 
\State $K$: the step length of the connection between nodes;
\State \textit{X}: the input feature map with the shape $H \times W$; 
\State \textit{X$_{quadrants}$}: the quadrants of the input flipped across the diagonals;
\State $m$: the distance of each roll.

\end{algorithmic}
\begin{algorithmic}[1]
\State $\textit{X}_{norm}$ $\leftarrow$ $\mathtt{norm}$(\textit{X}, \textit{X$_{quadrants}$}) \quad\quad $\triangleright$ \textcolor{magenta}{matrix norm of tensors}
\State $\mu$ $\leftarrow$ $\mathtt{mean}$($\textit{X}_{norm}$), \quad $\sigma$ $\leftarrow$ $\mathtt{std}$($\textit{X}_{norm}$)

\State \textbf{while} $mK$ $<$ \textit{H} \textbf{do}
\State \quad \quad \textit{X$_{rolled}$} $\leftarrow$ $\mathtt{roll}_{down}$(\textit{X}, $mK$), \quad \textit{dist} $\leftarrow$ $\mathtt{norm}$(\textit{X}, \textit{X$_{rolled}$}) \quad \quad $\triangleright$ \textcolor{magenta}{get distance value}
\State \quad \quad \textbf{if} \textit{dist} $<$ ($\mu$ $-$ $\sigma$) \textbf{then}  \quad \textit{mask} $\leftarrow$ 1 \quad $\triangleright$ \textcolor{magenta}{generate mask}
\State \quad \quad \textbf{else} \quad \textit{mask} $\leftarrow$ 0
\State \quad \quad \textbf{end if}
\State \quad \quad \textit{X$_{masked}$} $\leftarrow$ \textit{mask} $*$ (\textit{X$_{rolled}$} $-$ \textit{X})
\quad $\triangleright$ \textcolor{magenta}{get features}, \quad \textit{X$_{final}$} $\leftarrow$ $\mathtt{max}$(\textit{X$_{masked}$}, \textit{X$_{final}$})
\quad $\triangleright$ \textcolor{magenta}{keep max}
\State \quad \quad $m$ $\leftarrow$ $m+1$
\State \textbf{end while}

\State $m$ $\leftarrow$ 0
\State \textbf{while} $mK$ $<$ \textit{W} \textbf{do}
\State \quad \quad \textit{X$_{rolled}$} $\leftarrow$ $\mathtt{roll}_{right}$(\textit{X}, $mK$), \quad \textit{dist} $\leftarrow$ $\mathtt{norm}$(\textit{X}, \textit{X$_{rolled}$})
\State \quad \quad \textbf{if} \textit{dist} $<$ ($\mu$ $-$ $\sigma$) \textbf{then} \quad \textit{mask} $\leftarrow$ 1
\State \quad \quad \textbf{else} \quad \textit{mask} $\leftarrow$ 0
\State \quad \quad \textbf{end if}
\State \quad \quad \textit{X$_{masked}$} $\leftarrow$ \textit{mask} $*$ (\textit{X$_{rolled}$} $-$ \textit{X}), \quad \textit{X$_{final}$} $\leftarrow$ $\mathtt{max}$(\textit{X$_{masked}$}, \textit{X$_{final}$})
\State \quad \quad $m$ $\leftarrow$ $m+1$
\State \textbf{end while}
\State return $\mathtt{Conv2d}$($\mathtt{Concat}$(\textit{X}, \textit{X}$_{final}$))
\end{algorithmic}
\end{algorithm}

Now that we have obtained the estimated mean \( \mu \) and standard deviation \( \sigma \) of the image, we translate the input feature map \( X \) by \( m_K \) pixels either horizontally or vertically, provided that \( m_K \) is smaller than the height \( H \) and width \( W \) of the feature map, as shown in Algorithm 1. This translation operation is used to compare the feature patches that are separated by \( N \) steps. In Fig. \ref{fig:dynamic_graph_construction} (b), node $(5,1)$ in the coordinates of $(x,y)$ is compared to nodes $(3,2)$, $(6,2)$, $(7,2)$, $(5,3)$ through "rolling" to the next node. After the rolling operation, we compute the Euclidean distance between the input \(X\) and its rolled version \(X_{\text{rolled}}\) to determine whether the two points should be connected. If the distance is less than \(\mu - \sigma\), the mask is assigned a value of 1; otherwise, it is assigned a value of 0. This mask is then multiplied by \(X_{\textit{rolled}} - X\) to suppress the maximum relative scores between the feature patches that are not considered connected. In Algorithm \ref{alg:dynamic_graph}, these values are denoted as \(X_{\textit{down}}\) and \(X_{\textit{right}}\), respectively. Next, a max operation is performed, and the result is stored in \(X_{\textit{final}}\). Finally, after the rolling, masking, and max-relative operations, a final \( \mathtt{Conv2d} \) computation is applied.

Through our proposed approach, DGGF constructs a more representative graph structure compared to GGF \cite{guan2024talk2radar}, as it does not connect dissimilar patches (i.e., nodes). Additionally, compared to KNN, DGGF significantly reduces the computational overhead by minimizing adjacency computations during graph construction (whereas KNN must determine the nearest neighbors for each image patch). Furthermore, DGGF does not require the reshaping necessary for performing graph convolution in KNN-based methods. Therefore, DGGF combines the representational flexibility of KNN with the computational efficiency of GGF.

Based on the dynamically constructed graph described above, we connect it to a feedforward neural network with a GeLU activation. Given an input feature $f_{t}^{lc} \in \mathbb{R}^{N \times N}$, the detailed process of updated dynamic grapher is as follows:

\begin{equation}
    lc = \phi(\mathtt{DynConv}(f_{t}^{lc}))\mathbf{W}_{in})\mathbf{W}_{out} + lr,
    \label{eq:dyn_graph}
\end{equation}
where $lc \in \mathbb{R}^{N \times N}$ denotes the output graph. $\mathbf{W}_{in}$ and $\mathbf{W}_{out}$ are two weights of the feedforward layers. $\phi$ denotes the GeLU activation.


\begin{figure}
\centering
    \includegraphics[width=0.99\linewidth]{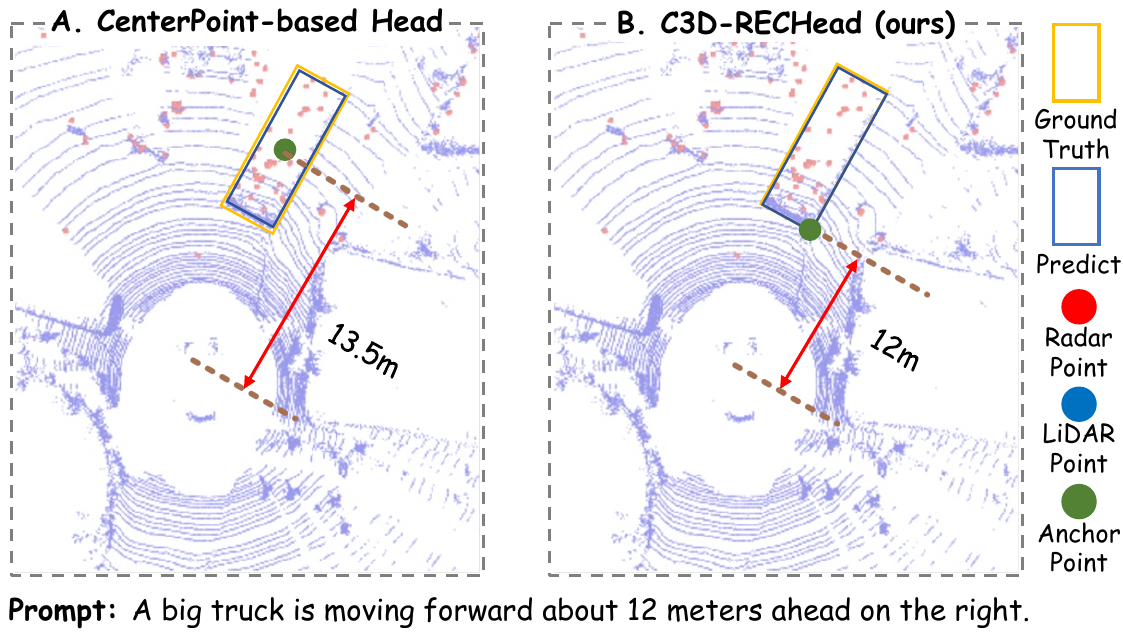}
    \vspace{-6mm}
    \caption{The difference between CenterPoint-based prediction head and our proposed C3D-RECHead.}
    \label{fig:c3d_head_vis}
\end{figure}

\subsection{C3D-RECHead}
\label{subsec:c3d}
Conventional 3D object detection models based on point clouds, such as CenterPoint \cite{yin2021center}, typically regress the bounding box relative to the object's center. However, in 3D visual grounding based on point cloud sensors, we identify two key limitations of the center-based regression: \textbf{(1) Higher Point Cloud Density at the Nearest Edge:} The nearest object to the autonomous vehicle has a denser point cloud, especially on the side facing the vehicle, which provides a more reliable anchor point for regression, improving localization accuracy. \textbf{(2) Alignment with Textual Distance Prompts:} Many textual prompts in 3D visual grounding include object distances (e.g., ``the car 5 meters ahead"). These distances typically refer to the edge of the object closest to the vehicle, rather than the center. Thus, regressing from this nearest edge point aligns better with semantic and perceptual grounding.

Instead of regressing from the object’s center, we propose C3D-RECHead (Fig. \ref{fig:c3d_head_vis}). Exactly, it performs bounding box regression from the nearest edge point to the vehicle. This adaptation preserves the computational efficiency of CenterPoint while improving localization accuracy in 3D visual grounding tasks. Assume a 3D bounding box $B$ is represented as:

\begin{equation}
    B = \{p_c, l, w, h, \theta\},
\end{equation}
where $p_c$ denotes the center of the box, $l, w, h$ denote length, width and height of the box, respectively, and $\theta$ denotes the orientation of the box.

For each bounding box, the eight corner points are:
\begin{equation}
    P = \{ p_1, p_2, \dots , p_8\},
    \label{eq:corner_points}
\end{equation}
where each $p_i$ is computed using the box dimensions and orientation.

Given the sensor position $p_s$, the edge set $E$ consists of four candidate edges (assuming a rectangular box base in BEV):
\begin{equation}
    E = \{e_1, e_2, e_3, e_4\},
    \label{eq:edges}
\end{equation}
where each edge is defined as two points:

\begin{equation}
    e_i = (p_i, p_j), i \neq j.
\end{equation}

We select the nearest edge $e^*$ by:

\begin{equation}
    e^* = \underset{e_i \in E}{\mathtt{argmin}} \ d(e_i, p_s),
\end{equation}
where $d(e_i, p_s)$ is the Euclidean distance from the sensor to each edge.

\subsection{Loss Function}
\label{subsec:loss}
The proposed C3D-RECHead first generates heatmaps to estimate the nearest corner of each detected object, which serves as an anchor point for subsequent regression. This design exploits the higher point density from LiDAR and radar near the closest object surface, thereby enhancing localization accuracy. Conditioned on the heatmap localization, the head refines object attributes following a CenterPoint-inspired supervision scheme. Specifically, it predicts: (1) sub-voxel offsets for localization precision beyond the voxel resolution; (2) height above ground to capture vertical positioning critical for 3D discrimination; (3) 3D bounding box dimensions \((w, l, h)\) to describe object geometry; and (4) orientation to model rotational pose within the 3D scene.

Finally, our proposed TPCNet is trained using a multi-task loss function, integrating these regression objectives to optimize performance in 3D visual grounding. The loss formulation is as follows:  

\begin{equation}
    {L}_{\rm total} = {L}_{\rm hm} + \beta \textstyle\sum\limits_{r \in \boldsymbol{\Lambda} } {L}_{{\rm smooth}-\ell_1}(\widehat{\Delta r^{a}}, \Delta r^{a}),
\end{equation}
where ${L}_{\rm hm}$ is the confidence loss supervising the heatmap quality of the center-based detection head using a focal loss; $\boldsymbol{\Lambda} = \left\{ x, y, z, l, h, w, \theta \right\}$ indicates the smooth-$\ell_1$ loss supervising the regression of the box center (for slight modification based on heatmap peak guidance), dimensions, and orientation; and $\beta$ is the weight to balance the two components of the loss, which is set to 0.25 by default.

\section{Experiments and Performance Analysis}
\label{sec:experiments}

\begin{table*}
    \setlength\tabcolsep{2.8pt}
    \caption{Overall performances on Talk2Radar dataset \cite{guan2024talk2radar} (\textbf{Best Performance}. \textbf{Car}, \textbf{Pedestrian} and \textbf{Cyclist} provide specialized \textbf{mAPs}. R5 denotes the accumulation of five frame radar data; C denotes camera while L denotes LiDAR.)}
    \vspace{-3mm}
    \label{tab:benchmark_compare}
    \centering
    \begin{tabular}{cc|ccc|cccc|c|cccc|c}
    \toprule
       \multirow{2}[2]{*}{\textbf{Models}} & \multirow{2}[2]{*}{\textbf{Venues}} & \multirow{2}[2]{*}{\textbf{Sensors}} & \multirow{2}[2]{*}{\textbf{Text Encoder}} & \multirow{2}[2]{*}{\textbf{Fusion}} & \multicolumn{5}{c}{\textbf{Entire Annotated Area (EAA)}}  & \multicolumn{5}{c}{\textbf{Driving Corridor Area (DCA)}}  \\
       \cmidrule(lr){6-10}  \cmidrule(lr){11-15}
        & & & & & \textbf{Car} & \textbf{Pedestrian} & \textbf{Cyclist} & \textbf{mAP} & \textbf{mAOS} & \textbf{Car} & \textbf{Pedestrian} & \textbf{Cyclist} & \textbf{mAP} & \textbf{mAOS}  \\
    \midrule
      PointPillars & CVPR$_{\text{2019}}$ &  R5 & ALBERT \cite{lan2019albert} & HDP & 18.92 & 9.79 & 12.47 & 13.73 & 12.91 &  39.20 & 10.25 & 14.93 & 21.46 & 20.19   \\
      CenterFormer & ECCV$_{\text{2022}}$ & R5 & ALBERT & HDP & 17.26 & 6.79 & 9.27 & 11.11 & 10.79 &  19.56 & 9.13 & 12.03 & 13.57 & 13.02 \\
      CenterPoint & ICCV$_{\text{2021}}$ & R5 & ALBERT & HDP  & 18.98 & 5.30 & 14.96 & 13.08 & 12.20 &  40.53 & 8.57 & 15.66 & 21.59 & 20.25  \\
      PointPillars & CVPR$_{\text{2019}}$ & R5 & ALBERT & MHCA & 5.18 & 5.76 & 6.63 & 5.86 & 3.58 & 13.34 & 4.36 & 8.79 & 8.83 & 7.66  \\
      CenterFormer & ECCV$_{\text{2022}}$ & R5 & ALBERT & MHCA & 4.53 & 3.48 & 4.00 & 4.00 & 2.03  & 8.77 & 3.52 & 6.69 & 6.33 & 5.92 \\
      CenterPoint & ICCV$_{\text{2021}}$ & R5 & ALBERT & MHCA & 5.21 & 4.57 & 5.13 & 4.97 & 3.12 & 12.70 & 4.07 & 7.70 & 8.16 & 7.51 \\
     MSSG \cite{cheng2023language} & arXiv$_{\text{2023}}$ & R5 & GRU \cite{chung2014empirical} & - & 12.53 & 5.08 & 8.47 & 8.69 & 7.03 & 18.93 & 7.88 & 9.40 & 12.07 & 11.67\\
     AFMNet \cite{solgi2024transformer} & AISP$_{\text{2024}}$ & R5 & GRU & - & 11.98 & 6.87 & 9.16 & 9.34 & 7.72 & 18.62 & 8.21 & 10.06 & 12.30 & 11.79 \\
     MSSG \cite{cheng2023language} & arXiv$_{\text{2023}}$ & R5 & ALBERT & - & 16.03 & 5.86 & 10.57 & 10.82 & 8.96 & 25.79 & 8.69 & 12.55 & 15.68 & 14.12 \\
     AFMNet \cite{solgi2024transformer} & AISP$_{\text{2024}}$ & R5 & ALBERT & - & 16.31 & 6.80 & 10.35 & 11.15 & 9.46 & 26.82 & 8.71 & 12.45 & 15.99 & 14.18 \\
     EDA \cite{wu2023eda} & CVPR$_{\text{2023}}$ & R5 & RoBERTa \cite{liu2019roberta} & - & 13.23 & 6.60 & 8.63 & 9.49 & 8.93 & 23.55 & 8.80 & 11.95 & 14.77 & 13.07 \\
    T-RadarNet \cite{guan2024talk2radar} & ICRA$_{\text{2025}}$ & R5 & ALBERT & GGF & 24.68 & 9.71 & 15.74 & 16.71 & 14.88 & 42.58 & 10.13 & 17.82 & 23.51 & 22.37  \\
    \textbf{TPCNet (ours)} & 2025 & R5 & PointCLIP \cite{zhang2022pointclip} & DGGF & 25.92 & 9.60 & 16.37 & 17.30 & 15.29 & 45.97 & 12.73 & 19.73 & 26.14 & 24.73\\
    \midrule
      CenterPoint & ICCV$_{\text{2021}}$ & L & ALBERT & HDP & 28.16 & 6.21 & 17.46 & 17.28 & 16.03 & 43.43 & 6.87 & 27.18 & 25.83 & 24.93   \\
      CenterPoint & ICCV$_{\text{2021}}$ & L & ALBERT & MHCA & 6.56 & 5.04 & 5.33 & 5.64 & 4.86 &  13.60 & 4.52 & 7.32 & 8.48 & 7.89 \\
     MSSG \cite{cheng2023language} & arXiv$_{\text{2023}}$ & L & GRU &  - & 15.38 & 7.52 & 11.67 & 11.52 & 9.76 & 23.27 & 8.68 & 13.51 & 15.15 & 14.75\\
     AFMNet \cite{solgi2024transformer} & AISP$_{\text{2024}}$ & L & GRU &  - & 16.13 & 7.68 & 12.51 & 12.11 & 9.92 & 24.50 & 9.07 & 13.87 & 15.81 & 15.11 \\
     MSSG \cite{cheng2023language} & arXiv$_{\text{2023}}$ & L & ALBERT &  - &  18.19 & 7.66 & 11.63 & 12.49 & 10.91 & 29.61 & 10.98 & 14.66 & 18.42 & 16.23 \\
     AFMNet \cite{solgi2024transformer} & AISP$_{\text{2024}}$ & L & ALBERT &  - & 19.50 & 7.92 & 13.56 & 13.66 & 12.18 & 31.68 & 9.23 & 18.90 & 19.94 & 17.59 \\
     EDA \cite{wu2023eda} & CVPR$_{\text{2023}}$ & L & RoBERTa & - & 16.10 & 6.91 & 12.88 & 11.96 & 10.10 & 25.10 & 9.28 & 15.73 & 16.70 & 14.91 \\
     T-RadarNet \cite{guan2024talk2radar} & ICRA$_{\text{2025}}$ & L & ALBERT & GGF & 24.91 & 12.74 & 18.67 & 18.77 & 17.20 & 48.98 & 14.69 & 27.24 & 30.30 & 29.89 \\
     \textbf{TPCNet (ours)} & 2025 & L & PointCLIP & DGGF & 28.93 & 12.83 & 20.72 & 20.83 & 18.94 & 49.78 & 15.25 & 29.11 & 31.38 & 30.14 \\
     \midrule
     BEVFusion \cite{liu2023bevfusion} & ICRA$_{\text{2023}}$ & C + R5 & PointCLIP & DGGF & 26.93 & 10.20 & 16.62 & 17.92 & 15.77 & 46.17 & 12.18 & 21.87 & 26.74 & 26.02 \\
     BEVFusion \cite{liu2023bevfusion} & ICRA$_{\text{2023}}$ & C + L & PointCLIP & DGGF & 29.07 & 14.45 & 20.50 & 21.34 & 19.72 & 50.82 & 15.65 & 29.37 & 31.95 & 30.41 \\
     \textbf{TPCNet (ours)} & 2025 & L + R5 & PointCLIP & DGGF & \textbf{32.13} & \textbf{15.99} & \textbf{23.74} & \textbf{23.95} & \textbf{22.01} & \textbf{52.72} & \textbf{17.30} & \textbf{31.66} & \textbf{33.89} & \textbf{31.73} \\
    \bottomrule
    \end{tabular}
\end{table*}

\subsection{Settings of Models and Implementations}
\textbf{Model Settings:} Besides the PointCLIP \cite{zhang2022pointclip} that we leverge in our proposed TPCNet, for other models with pre-trained transformer (e.g., ALBERT \cite{lan2019albert}) for text encoding, we set the token length as 30, uniformly. The number of pillars in the pillar backbone of TPCNet is set as 10 and 32 for radar and LiDAR, respectively. We set the sizes of both LiDAR and Radar Context Agent Feature as $12 \times 12$ in default. For the $K$ values of graph construction in DGGF, we set 8, 4, 2 for three stage feature maps, respectively.

For comparison, we \textbf{firstly} select PC-based detectors with various paradigms including PointPillars (pillar-based) \cite{lang2019pointpillars}, CenterPoint (voxel and anchor-free) \cite{yin2021center}, and CenterFormer (transformer-based) \cite{zhou2022centerformer}, which all fuse the point cloud and textual features between the stage of backbone and FPN, following the same fusion paradigm with our proposed TPCNet. Among the above models, we leverage the SECOND FPN \cite{yan2018second} as the multi-scale feature fusion module in PointPillars and CenterPoint. \textbf{Secondly}, the fusion methods of point clouds and text include inductive bias-based HDP \cite{zhu2022seqtr}, attention-based MHCA \cite{wu2023referring} and graph-based GGF \cite{guan2024talk2radar} are implemented to compare with our DGGF. \textbf{Lastly}, we also compare proposed TPCNet with dedicated SOTA 3D PC grounding models, including T-RadarNet \cite{guan2024talk2radar}, MSSG \cite{cheng2023language}, AFMNet \cite{solgi2024transformer} and EDA \cite{wu2023eda}. Moreover, various LiDAR and radar fusion approaches, including Multi-Head Cross Attention (MHCA) \cite{vaswani2017attention} and Multi-Head Linear Attention (MHLCA) \cite{choromanski2020rethinking}, InterRAL \cite{wang2022interfusion}, L2R Fusion \cite{wang2023bi} and Adaptive Gated Network (AGN) \cite{song2024lirafusion}, are compared with our fusion approach as well.

\textbf{Dataset Settings:} We conduct comprehensive training and evaluation of models on the Talk2Radar dataset \cite{guan2024talk2radar}, which encompasses three distinct categories of traffic participants: \textit{Car}, \textit{Cyclist}, and \textit{Pedestrian}, utilizing both 4D radar and LiDAR modalities. Furthermore, to rigorously assess the generalization capabilities of T-RadarNet, we extend our evaluation to the Talk2Car dataset \cite{deruyttere2019talk2car}, which provides synchronized LiDAR and radar data accompanied by textual prompts referencing object objects, thereby enabling a robust validation of our approach across diverse multi-modal scenarios.

\begin{table}
\setlength\tabcolsep{1.2pt}
\footnotesize
    \caption{Performances on Talk2Car for 3D REC, including results in adverse environments. The \textbf{AP$_\text{A}$} and \textbf{AP$_\text{B}$} follow MSSG \cite{cheng2023language} that define different IoU thresholds. L denotes LiDAR while R denotes Radar.}
    \vspace{-3mm}
    \label{tab:talk2car_compare}
    \centering
    \begin{tabular}{c|c|cccc|cc}
    \toprule
    \multirow{2}[2]{*}{\textbf{Models}} & \multirow{2}[2]{*}{\textbf{Sensors}} & \multicolumn{2}{c}{\textbf{BEV AP}} & \multicolumn{2}{c}{\textbf{3D AP}} & \textbf{Rain} & \textbf{Night}  \\
    \cmidrule{3-4} \cmidrule{5-6} \cmidrule{7-8}
         & & \textbf{AP$_\text{A}$} \ $\uparrow$ & \textbf{AP$_\text{B}$} \ $\uparrow$ & \textbf{AP$_\text{A}$} \ $\uparrow$ & \textbf{AP$_\text{B}$} \ $\uparrow$ & \textbf{AP$_\text{A}$} \ $\uparrow$ & \textbf{AP$_\text{A}$} \ $\uparrow$ \\
    \midrule
    Baseline \cite{deruyttere2019talk2car} & L & 30.6 & 24.4 & 27.9 & 19.1 & - & - \\
    MSSG \cite{cheng2023language} & L & 27.8 & 26.1 & 31.9 & 20.3 & - & -\\
    EDA \cite{wu2023eda} & L & 37.0 & 29.8 & 37.2 & 20.4 & 18.9 & 13.2 \\
    AFMNet \cite{solgi2024transformer} & L & 45.3 & 33.1 & 41.9 & 20.7 & 23.8 & 15.0\\
    T-RadarNet \cite{guan2024talk2radar} & L & 52.8 & 39.9 & 47.2 & 30.5 & 24.3 & 17.4 \\
    \midrule
    \textbf{TPCNet (ours)} & L & 53.3 & 42.7 & 46.8 & 31.6 & 24.9 & 19.0 \\
    \textbf{TPCNet (ours)} & L+R & \textbf{56.7} & \textbf{44.1} & \textbf{52.3} & \textbf{33.6} & \textbf{27.5} & \textbf{21.8} \\
    \bottomrule
    \end{tabular}
\end{table}

\textbf{Training and Evaluation Settings:} For the Talk2Radar dataset, all models are trained using a distributed setup across four RTX A4000 GPUs, with a batch size of 4 per GPU for a total of 80 epochs. The optimization process employs AdamW with an initial learning rate of \(1 \times 10^{-3}\), modulated by a cosine annealing learning rate scheduler, and a weight decay of \(5 \times 10^{-4}\). To evaluate 3D visual grounding performance, we utilize two key metrics: Average Precision (AP) and Average Orientation Similarity (AOS), which are computed for both the entire annotated area and the driving corridor. Specifically, we report the mean Average Precision (mAP) and mean Average Orientation Similarity (mAOS) for 3D bounding box localization and orientation estimation.

For the Talk2Car dataset, we adhere to the baseline configurations outlined in \cite{cheng2023language}. Specifically, models are trained on four RTX A4000 GPUs with a batch size of 1 per GPU for 20 epochs. Optimization is performed using Stochastic Gradient Descent (SGD) with a momentum of 0.9, a weight decay of \(1 \times 10^{-4}\), and an initial learning rate of \(1 \times 10^{-2}\), which is similarly adjusted via a cosine annealing scheduler. The evaluation metric for 3D visual grounding is exclusively based on Average Precision (AP), which quantifies the localization accuracy of the predicted 3D bounding boxes relative to the ground truth annotations.

\subsection{Comparison with State-of-the-arts}
\textbf{Overall comparison with state-of-the-art approaches on Talk2Radar and Talk2Car datasets.} 
As shown in Table \ref{tab:benchmark_compare}, our proposed TPCNet, which integrates LiDAR and radar, achieves state-of-the-art performance on the Talk2Radar dataset. Compared to unimodal models (LiDAR-only and radar-only), TPCNet demonstrates a significant improvement in both mAP and mAOS metrics. Furthermore, even in its unimodal configuration, TPCNet outperforms the models with similar sensor inputs, highlighting the superiority of our model architecture. In terms of per-class accuracy, the detection performance for Car is slightly higher than that for Cyclist, while Pedestrian exhibits the lowest accuracy. Additionally, on the Talk2Car dataset (Table \ref{tab:talk2car_compare}), TPCNet also surpasses the best-performing LiDAR-only model, demonstrating strong generalization across sensor data distributions.

\begin{table}
\setlength\tabcolsep{2.5pt}
    \caption{Comparison (\textbf{mAP}) of different modalities upon various prompts in the Talk2Radar dataset \cite{guan2024talk2radar}. R1, R3 and R5 denote the accumulation of one, three and five frames of radar data, respectively. L denotes LiDAR. The experiments of single modality are implemented by T-RadarNet \cite{guan2024talk2radar} and the performances of multiple modalities are implemented by TPCNet.}
    \vspace{-3mm}
    \label{tab:prompt_compare}
    \centering
    \begin{tabular}{c|ccc|ccc|ccc}
    \toprule
      \textbf{Prompt} & \multicolumn{3}{c}{\textbf{Motion}} & \multicolumn{3}{c}{\textbf{Depth}} & \multicolumn{3}{c}{\textbf{Velocity}}  \\
    \midrule
       \textbf{Sensors}  & \textbf{Car} & \textbf{Pedes} & \textbf{Cyclist} & \textbf{Car} & \textbf{Pedes} & \textbf{Cyclist} & \textbf{Car} & \textbf{Pedes} & \textbf{Cyclist}  \\
    \midrule
     R1 & 26.1 & 5.3 & 11.1 & 32.7 & 12.7 & 24.7 & 26.9 & 14.0 & 18.5  \\
     R3 & 35.9 & 11.1 & 15.4 & 40.7 & 18.7 & 31.7 & 33.8 & 20.5 & 25.8\\
     R5 & 36.7 & 11.5 & 16.3 & 42.5 & 19.6 & 32.0 & 35.5 & 20.8 & 25.7\\
     L & 33.6 & 8.8 & 13.6 & 45.7 & 20.5 & 38.6 & 12.6 & 7.6 & 8.2\\
     \midrule
     \textbf{R1+L} & 37.6 & 13.4 & 18.9 & 48.8 & 22.5 & 41.9 & 38.7 & 26.5 & 29.8 \\
     \textbf{R3+L} & 37.7 & 12.8 & 19.7 & 48.2 & 24.1 & 42.5 & 39.6 & 27.8 & 30.9\\
     \textbf{R5+L} & \textbf{38.9} & \textbf{14.1} & \textbf{21.2} & \textbf{49.7} & \textbf{25.7} & \textbf{42.7} & \textbf{40.8} & \textbf{28.6} & \textbf{32.0} \\
    \bottomrule     
    \end{tabular}  
\end{table}

\begin{table}
\setlength\tabcolsep{4.0pt}
    \caption{Statistics of \textbf{mAP} for predicted objects by TPCNet with the respect with depth upon 5-frame radar, LiDAR and the combination of radar and LiDAR.}
    \vspace{-3mm}
    \label{tab:distance_compare}
    \centering
    \begin{tabular}{c|c|cccccc}
    \toprule
      \textbf{Sensors} & \textbf{Objects} & \textbf{0-10 (m)} & \textbf{10-20} & \textbf{20-30} & \textbf{30-40} & \textbf{40-50} & \textbf{50+}  \\
    \midrule
     \multirow{3}[3]{*}{R5} &  Car & 43.62 & 45.19 & 25.32 & 20.13 & 11.28 & 2.20 \\
      & Pedestrian & 16.79 & 8.11 & 3.98 & 3.09 & 4.89 & 0.0  \\
      &  Cyclist & 31.55 & 14.23 & 16.17 & 5.74 & 3.23 & 1.22  \\
    \midrule
    \multirow{3}[3]{*}{L} & Car & 46.78 & 49.50 & 31.72 & 26.45 & 14.19 & 8.12 \\
      & Pedestrian & 19.55 & 9.79 & 7.88 & 6.57 & 5.48 & 0.0 \\
      &  Cyclist & 36.92 & 17.63 & 20.11 & 7.53 & 6.11 & 3.74 \\
    \midrule
    \textbf{L} & Car & 51.67 & 54.69 & 36.55 & 29.87 & 18.56 & 10.59 \\
    \textbf{+} & Pedestrian & 21.41 & 11.16 & 7.92 & 7.10 & 5.79 & 0.0 \\
    \textbf{R5} & Cyclist & 38.55 & 21.27 & 22.79 & 10.57 & 6.20 & 4.12 \\
    \bottomrule
    \end{tabular}
\end{table}

\textbf{Performances of TPNet based on two input sensors for different types of prompts.} Table \ref{tab:prompt_compare} presents the performance of TPCNet with different sensor inputs when various types of prompts are processed. Overall, for text prompts containing motion or velocity information, radar generally outperforms LiDAR. In contrast, for prompts involving object depth, the LiDAR-only model performs better than the radar-only model. However, the fusion of radar and LiDAR significantly outperforms both uni-modal approaches. Additionally, aggregating five frames of radar point clouds yields better performance than using three or one frame. This demonstrates the strong complementarity between radar and LiDAR.

\begin{table}
\setlength\tabcolsep{4.0pt}
\footnotesize
    \caption{Performances of fusion methods for LiDAR and radar inputs in TPCNet on Talk2Radar \cite{guan2024talk2radar} and Talk2Car \cite{deruyttere2019talk2car} dataset. The \textbf{AP$_{\text{A}}$} and \textbf{AP$_{\text{B}}$} denotes the 3D AP.}
    \vspace{-3mm}
    \label{tab:radar_lidar_fusion_compare}
    \centering
    \begin{tabular}{cc|cc|cc}
    \toprule
      \multirow{2}[2]{*}{\textbf{Methods}} & \multirow{2}[2]{*}{\textbf{Venues}} & \multicolumn{2}{c}{\textbf{Talk2Radar}} & \multicolumn{2}{c}{\textbf{Talk2Car}}  \\
       \cmidrule(lr){3-4} \cmidrule(lr){5-6}
         & & \textbf{mAP} \ $\uparrow$ & \textbf{mAOS} \ $\uparrow$ & \textbf{AP$_\text{A}$} \ $\uparrow$ & \textbf{AP$_\text{B}$} \ $\uparrow$ \\
    \midrule
     Bi-MHCA \cite{wu2023referring} & CVPR$_{\text{2023}}$ & \textbf{29.14} & 25.87 & 49.7 & 33.2 \\
     Bi-MHLCA \cite{choromanski2020rethinking} & ICLR$_{\text{2020}}$ & 24.53 & 23.92 & 48.1 & 30.9\\
     InterRAL \cite{wang2022interfusion} & IROS$_{\text{2022}}$ & 26.45 & 24.23 & 50.8 & 31.3\\
     L2R Fusion \cite{wang2023bi} & CVPR$_{\text{2023}}$ & 25.26 & 22.94 & 48.5 & 30.1 \\
     AGN \cite{song2024lirafusion} & ICRA$_{\text{2024}}$ & 27.84 & 25.03 & 50.1 & 31.0 \\
     \midrule
     \textbf{BACA (ours)} & 2025 & 28.92 & \textbf{26.07} & \textbf{52.3} & \textbf{33.6} \\
    \bottomrule     
    \end{tabular}
\end{table}

\textbf{Performances of various fusion methods of LiDAR and radar under the settings of TPCNet.} As TABLE \ref{tab:radar_lidar_fusion_compare} shows, under the TPCNet architecture, our BACA outperforms other LiDAR-radar fusion methods. Overall, it can be observed that the global fusion methods based on cross-attention achieve superior performance compared with convolution-based fusion approaches. Specifically, the Bi-MHCA-based fusion method achieves the highest mAP, followed closely by our proposed BACA, which performs slightly worse than Bi-MHCA but surpasses Bi-MHLCA. Moreover, on the Talk2Car dataset, our BACA achieves the best performance.

\begin{table}
    \setlength\tabcolsep{11.5pt}
    \caption{\textmd{Comparison of efficiency on fusion methods of LiDAR and radar features with different sizes. $N_h$ denotes the number of heads in cross attention, which is only for the attention-based fusion method. \textbf{FPS} is the frame per second of TPCNet equipped with various fusion methods on single RTX A4000 GPU.}}
    \vspace{-3mm}
    \label{tab:lidar_radar_efficiency_compare}
    \centering
    \begin{tabular}{c|cc|c}
    \toprule
        \textbf{Methods} & \textbf{Params}\ $\downarrow$ & \textbf{FLOPs}\ $\downarrow$ & \textbf{FPS}\ $\uparrow$ \\
    \midrule
    \multicolumn{4}{c}{\textcolor{green}{Stage 3 of pseudo image: }\textbf{$D$=64, $H$=80, $W$=80, $N_h$=4}} \\
    \multicolumn{4}{c}{\textcolor{blue}{Stage 4 of pseudo image: }\textbf{$D$=128, $H$=40, $W$=40, $N_h$=8}} \\
    \multicolumn{4}{c}{\textcolor{red}{Stage 5 of pseudo image: }\textbf{$D$=256, $H$=20, $W$=20, $N_h$=8 }} \\
    \midrule
    \multicolumn{4}{c}{\textbf{Non-Attention-based Fusion Methods}} \\
    \midrule
    \multirow{3}[3]{*}{AGN \cite{song2024lirafusion}} & \textcolor{green}{147.58K} & \textcolor{green}{1.89G}  \\
    & \textcolor{blue}{590.08K} & \textcolor{blue}{1.89G} & \textbf{19.7} \\
    & \textcolor{red}{2.36M} & \textcolor{red}{1.89G} \\
    \midrule
    \multicolumn{4}{c}{\textbf{Attention-based Fusion Methods}} \\
    \midrule
    \multirow{3}[3]{*}{Bi-MHCA \cite{vaswani2017attention}} & \textcolor{green}{0.33M}  & \textcolor{green}{10.70G}   \\
    & \textcolor{blue}{0.13M} & \textcolor{blue}{1520.44M} & 4.8 \\
    & \textcolor{red}{0.52M} & \textcolor{red}{373.56M} \\
    \midrule
    \multirow{3}[3]{*}{Bi-MHLCA \cite{choromanski2020rethinking}} & \textcolor{green}{131.14K} & \textcolor{green}{1.68G} \\
    & \textcolor{blue}{262.27K} & \textcolor{blue}{838.86M} & 9.5 \\
    & \textcolor{red}{524.54K} & \textcolor{red}{419.43M} \\
    \midrule
    \textbf{BACA (ours)} & \textcolor{green}{\textbf{0.64K}} & \textcolor{green}{\textbf{46.61M}} & \\
    \textbf{Agent Feature Len} & \textcolor{blue}{\textbf{1.28K}} & \textcolor{blue}{\textbf{23.33M}} & 18.2 \\
    \textbf{=8$\times$8} & \textcolor{red}{\textbf{2.56K}} & \textcolor{red}{\textbf{6.81M}} & \\
    \midrule
    \textbf{BACA (ours)} & \textcolor{green}{\textbf{0.64K}} & \textcolor{green}{\textbf{164.62M}} & \\
    \textbf{Agent Feature Len} & \textcolor{blue}{\textbf{1.28K}} & \textcolor{blue}{\textbf{82.40M}} & 16.2 \\
    \textbf{=16$\times$16} & \textcolor{red}{\textbf{2.56K}} & \textcolor{red}{\textbf{21.75M}} & \\
    \midrule
    \textbf{BACA (ours)} & \textcolor{green}{\textbf{0.64K}} & \textcolor{green}{\textbf{206.41M}} & \\
    \textbf{Agent Feature Len} & \textcolor{blue}{\textbf{1.28K}} & \textcolor{blue}{\textbf{103.32M}} & 13.0 \\
    \textbf{=18$\times$18} & \textcolor{red}{\textbf{2.56K}} & \textcolor{red}{\textbf{27.04M}} & \\
    
    \bottomrule
    \end{tabular}
\end{table}

\textbf{The comparison of efficiency between various fusion methods for LiDAR and radar.} As shown in Table \ref{tab:lidar_radar_efficiency_compare}, we have compared the complexity and inference speeds of non-attention-based fusion methods (convolution-based AGN) and attention-based fusion methods for processing multi-scale feature maps from the two modalities. We observe that our proposed BACA has significantly fewer parameters than other fusion methods, with a 2 to 3 orders of magnitude reduction in the parameter count when compared to the other three methods. Moreover, the FLOPs of BACA are substantially lower than those of Bi-MHLCA, Bi-MHCA, and AGN. At the same time, BACA achieves a much higher FPS than other attention-based fusion methods and is very close to the convolution-based AGN in terms of inference speed. Furthermore, when setting the agent feature map size to 8×8, 16×16, and 18×18, the increase in FLOPs remains relatively small, demonstrating the strong scalability of BACA.

\textbf{The comparison between CenterPoint-based prediction head and our proposed C3D-RECHead.} As shown in Table \ref{tab:head_compare}, we comprehensively evaluate the accuracy of 3D object detection and orientation angle estimation using two different detection head paradigms on the Talk2Radar dataset. Our findings indicate that the proposed C3D-RECHead exhibits a significant performance advantage over CenterPoint, achieving notable improvements in both mAP and mAOS.

\begin{table}
    \setlength\tabcolsep{1.5pt}
    \caption{Comparison of performances between C3D-RECHead and CenterPoint-based prediction head.}
    \vspace{-3mm}
    \label{tab:head_compare}
    \centering
    \begin{tabular}{c|cc|cc}
    \toprule
    \multirow{2}[2]{*}{\textbf{Methods}} & \multicolumn{2}{c}{\textbf{Entire Annotated Area}} & \multicolumn{2}{c}{\textbf{Driving Corridor Area}}  \\
       \cmidrule(lr){2-3} \cmidrule(lr){4-5}
         & \ \textbf{mAP} & \ \textbf{mAOS} & \ \textbf{mAP} & \ \textbf{mAOS} \\
    \midrule
        CenterPoint \cite{yin2021center} & 22.70 & 20.55 & 31.92 & 30.99 \\
    \midrule
        \textbf{C3D-RECHead (ours)} & \textbf{23.95} & \textbf{22.01} & \textbf{33.89} & \textbf{31.73} \\
    \bottomrule
    \end{tabular}
\end{table}

\begin{table}
\setlength\tabcolsep{2.5pt}
    \caption{Performances of fusion paradigm under our proposed BACA fusion method. R$_{\text{Q}}$+L$_{\text{KV}}$ and L$_{\text{Q}}$+R$_{\text{KV}}$ are unidirectional within the framework of BACA. R and L denote radar and LiDAR, respectively.}
    \vspace{-3mm}
    \label{tab:bi_fusion_compare}
    \centering
    \begin{tabular}{c|ccc|ccc|ccc}
    \toprule
      \textbf{Prompt} & \multicolumn{3}{c}{\textbf{Motion}} & \multicolumn{3}{c}{\textbf{Depth}} & \multicolumn{3}{c}{\textbf{Velocity}}  \\
    \midrule
       \textbf{Methods}  & \textbf{Car} & \textbf{Pedes} & \textbf{Cyclist} & \textbf{Car} & \textbf{Pedes} & \textbf{Cyclist} & \textbf{Car} & \textbf{Pedes} & \textbf{Cyclist}  \\
    \midrule
     R$_{\text{Q}}$+L$_{\text{KV}}$ & 37.9 & 10.9 & 19.8 & 45.5 & 21.2 & 38.1 & 38.9 & 26.9 & 29.7 \\
     L$_{\text{Q}}$+R$_{\text{KV}}$ & 36.2 & 9.1 & 17.6 & 49.0 & 24.8 & 40.9 & 36.0 & 22.0 & 24.9\\
     \midrule
     \textbf{BACA} & \textbf{38.9} & \textbf{14.1} & \textbf{21.2} & \textbf{49.7} & \textbf{25.7} & \textbf{42.7} & \textbf{40.8} & \textbf{28.6} & \textbf{32.0} \\
    \bottomrule     
    \end{tabular}  
\end{table}

\begin{figure*}
    \includegraphics[width=0.998\linewidth]{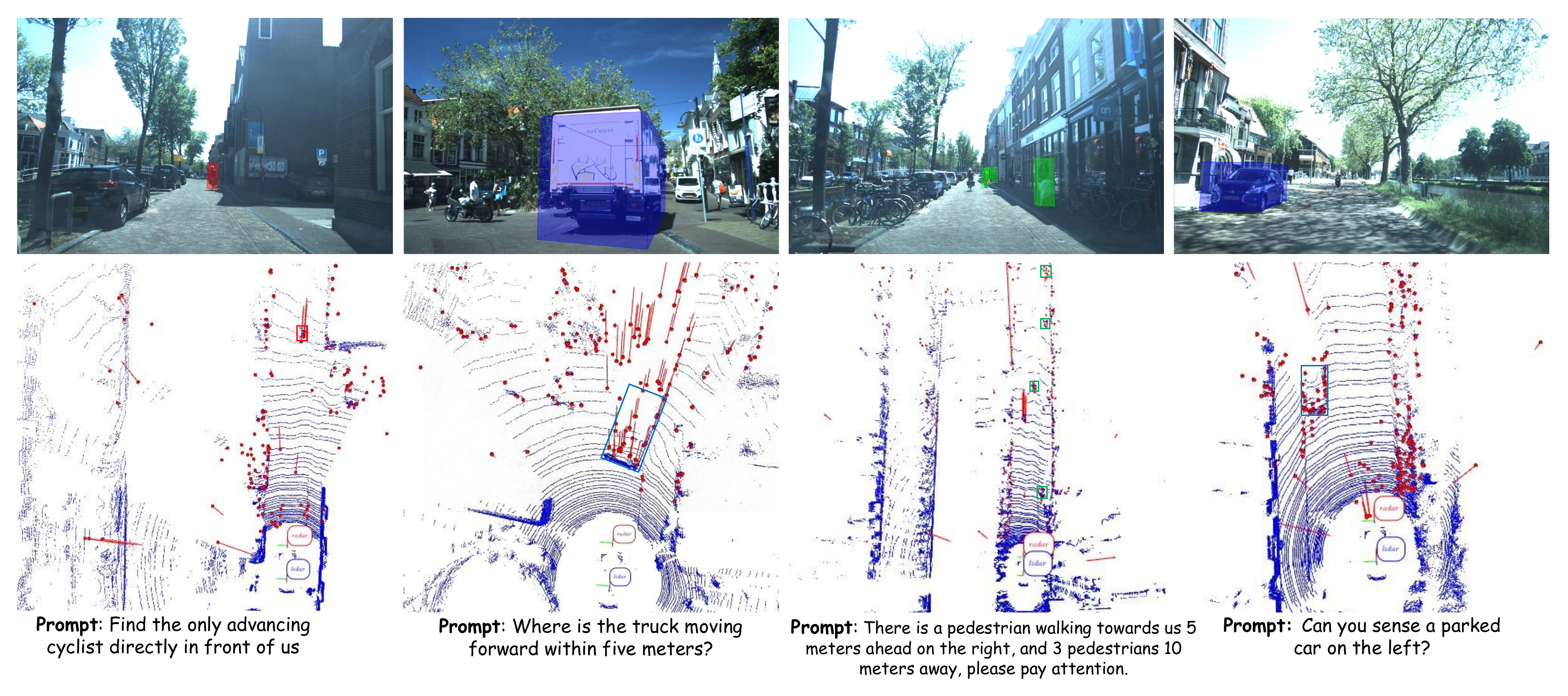}
    \vspace{-8mm}
    \caption{Visualization of predicted results by proposed TPCNet in Talk2Radar dataset. The first row presents the prediction from the view of camera while the second row shows the results from Bird's Eye View (BEV).}
    \label{fig:overall_visualization}
\end{figure*}

\begin{figure}
    \centering  
    \includegraphics[width=0.97\linewidth]{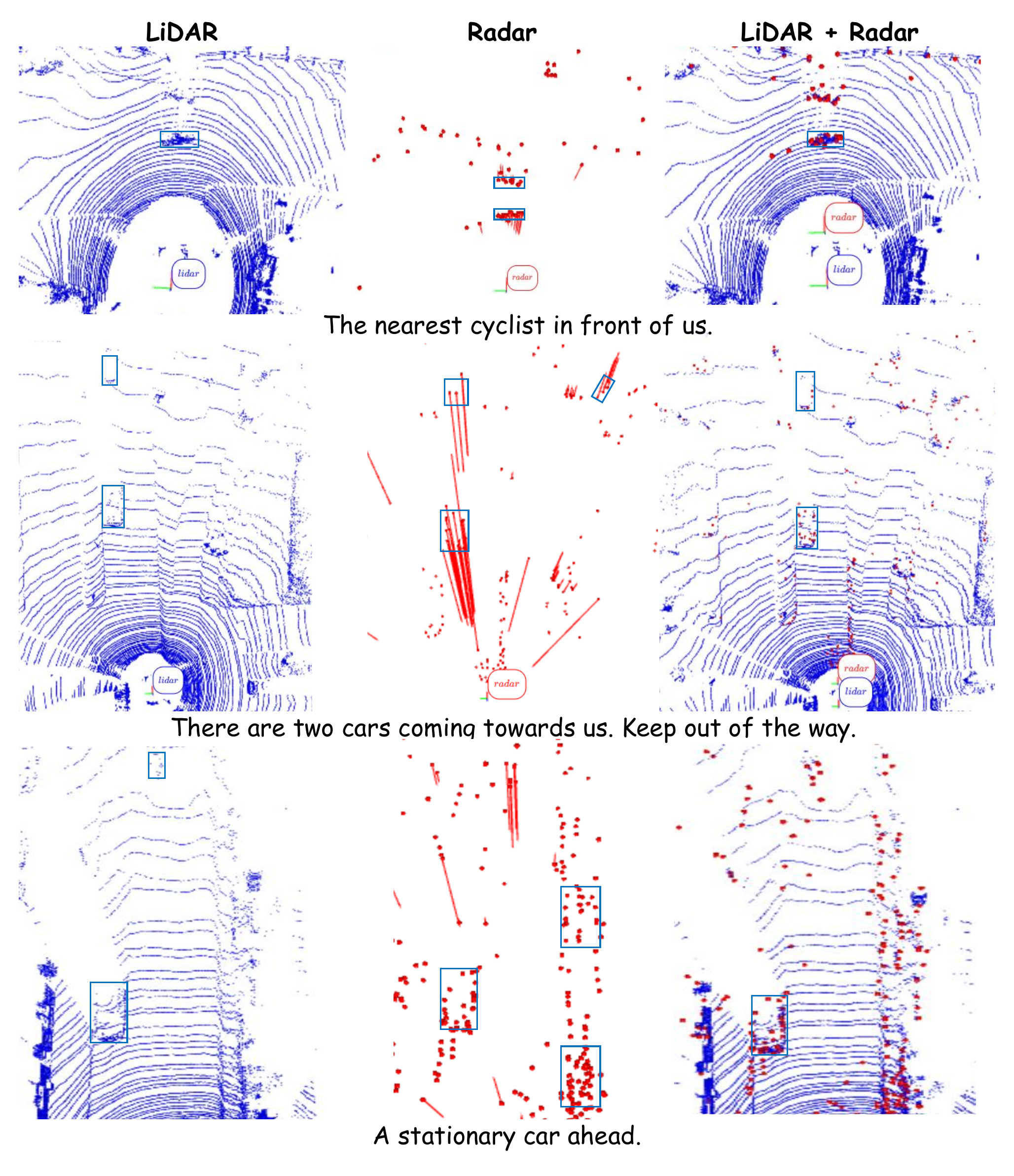}
    \vspace{-5mm}
    \caption{Prediction by LiDAR-only, radar-only and the fusion of LiDAR and radar based on the TPCNet.}
    \label{fig:lidar_radar_compare}
\end{figure}

\textbf{The comparison of the architecture of models.} As Fig. \ref{fig:lidar_radar_compare} shows, we have compared T-RadarNet and TPCNet, with the models taking radar and LiDAR as single-modal inputs, respectively, to analyze the effectiveness of the model architecture. Our results show that, regardless of whether radar or LiDAR is used as the input, TPCNet consistently outperforms T-RadarNet in accuracy across all object categories and at various distance intervals from the ego vehicle, demonstrating the architectural superiority of TPCNet over T-RadarNet.

\begin{figure}
\centering
    \includegraphics[width=0.99\linewidth]{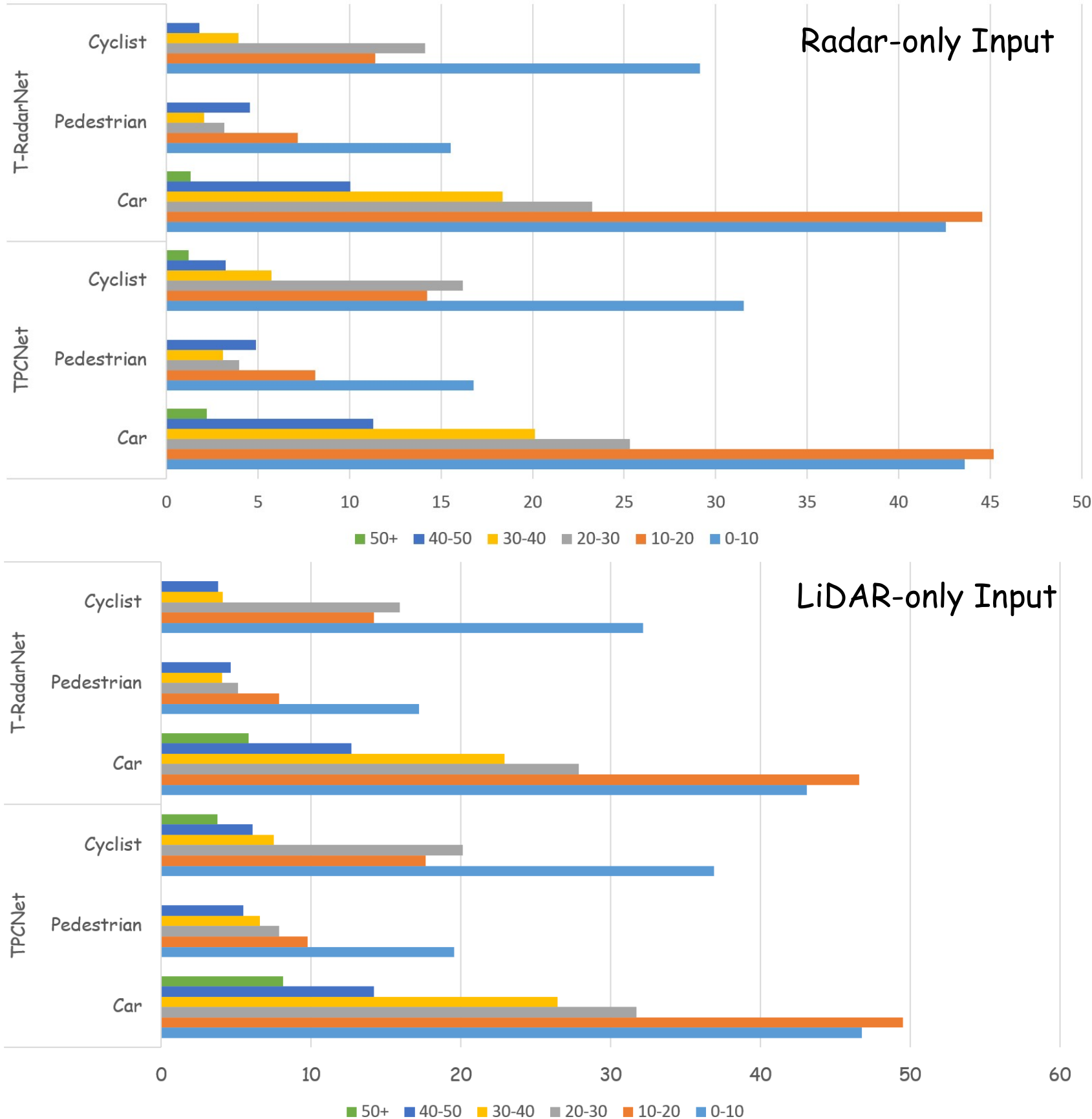}
    \vspace{-5mm}
    \caption{Comparison between TPCNet and T-RadarNet by depth based on the input of uni-modality.}
    \label{fig:depth_map_compare}
\end{figure}

\textbf{Statistics of prediction accuracy at different depths.} We partition the predicted objects into six depth intervals, each spanning 10 meters relative to the ego vehicle. As shown in Table \ref{tab:distance_compare} and Fig. \ref{fig:depth_map_compare}, the fusion of two sensors significantly enhances object detection accuracy, particularly for distant objects. The fused model outperforms both LiDAR-only and radar-only modalities, demonstrating the necessity of multi-modal fusion.

\begin{table}
\setlength\tabcolsep{4.6pt}
    \caption{Ablation experiments of TPCNet with various settings on Talk2Radar dataset.}
    \vspace{-3mm}
    \label{tab:ablation_exp}
    \centering
    \begin{tabular}{c|ccc|ccc}
    \hline
    \multirow{2}[2]{*}{\textbf{Methods}}  & \multicolumn{3}{c}{\textbf{EAA}} & \multicolumn{3}{c}{\textbf{DCA}} \\
     \cmidrule(lr){2-4} \cmidrule(lr){5-7}
     & \textbf{Car} & \textbf{Ped} & \textbf{Cyc} & \textbf{Car} & \textbf{Ped} & \textbf{Cyc} \\
    \toprule
      \multicolumn{7}{c}{\textbf{Dynamic Gated Graph Fusion}} \\
    \midrule
      \textbf{Dynamic} $\rightarrow$ Static \cite{munir2023mobilevig}  & 29.97 & 13.98 & 21.40 & 50.79 & 15.89 & 28.63 \\
      \textbf{MaxPool} $\rightarrow$ AvgPool & 30.19 & 14.67 & 20.55 & 51.41 & 16.36 & 29.74 \\
      \textbf{Dynamic + MaxPool} & \textbf{32.13} & \textbf{15.99} & \textbf{23.74} & \textbf{52.72} & \textbf{17.30} & \textbf{31.66}  \\
    \midrule
      \multicolumn{7}{c}{\textbf{Text Encoders}} \\
    \midrule
    Bi-GRU \cite{chung2014empirical} & 25.97 & 11.80 & 17.11 & 46.59 & 12.58 & 24.77 \\
    RoBERTa \cite{liu2019roberta} & 30.03 & 13.79 & 21.14 & 51.88 & 15.13 & 28.10 \\
    ALBERT \cite{lan2019albert} & 29.98 & 14.01 & 20.97 & 50.29 & 16.42 & 29.77 \\
    \textbf{PointCLIP} \cite{zhang2022pointclip} & \textbf{32.13} & \textbf{15.99} & \textbf{23.74} & \textbf{52.72} & \textbf{17.30} & \textbf{31.66} \\
    \midrule
     \multicolumn{7}{c}{\textbf{Fusion Methods of Text and Point Cloud}} \\
    \midrule
    HDP \cite{zhu2022seqtr} & 22.57 & 14.68 & 16.39 & 43.78 & 10.57 & 23.66 \\
    MHCA \cite{wu2023referring} & 11.23 & 6.54 & 8.96 & 18.20 & 7.96 & 10.29 \\
    MHLCA \cite{choromanski2020rethinking} & 9.12 & 6.74 & 7.05 & 16.46 & 8.41 & 10.10 \\
    GGF \cite{guan2024talk2radar} & 28.79 & 13.10 & 19.87 & 48.92 & 14.22 & 28.67 \\
    \textbf{DGGF} & \textbf{32.13} & \textbf{15.99} & \textbf{23.74} & \textbf{52.72} & \textbf{17.30} & \textbf{31.66} \\
    \bottomrule
    \end{tabular}
\end{table}

\subsection{Ablation Experiments}
\textbf{Ablation studies of vital modules in TPCNet.} As shown in TABLE \ref{tab:ablation_exp}:
(1) We first conduct an ablation study on the individual components of the DGGF module. Our analysis reveal that dynamic graphs achieve higher accuracy in scene object recognition when compared with static graphs, indicating that dynamic graphs handle environmental redundancy more effectively. Moreover, in constructing edges between graph nodes, MaxPool outperforms AvgPool in capturing salient inter-node relationships.

(2) We compare the performance of different text encoders on the 3D visual grounding task. Transformer-based encoders significantly outperform Bi-GRU, demonstrating the efficiency of the attention mechanisms in extracting meaningful textual information. However, PointCLIP, which leverages contrastive learning between point clouds and text, outperforms RoBERTa and ALBERT, highlighting its advantage in embedding textual features with point cloud-conditioned features, making it particularly suitable for point cloud-based visual grounding tasks.

(3) We replace different text-point cloud feature fusion modules in TPCNet. Our proposed DGGF outperforms both GGF and HDP, which rely on point-to-point fusion. Additionally, we observe that two global cross-attention-based modules, MHCA and MHLCA, perform poorly. This was primarily due to the excessive noise in the point cloud features (especially from radar), causing the model to overemphasize non-object regions and resulting in a high false positive rate.

\textbf{Comparison of performances on cross attention between LiDAR and radar.} TABLE \ref{tab:bi_fusion_compare} shows the results of ablation studies in the BACA module. We compared two model configurations: one where radar features were used as the query while LiDAR features served as the key and value, and the other where LiDAR was used as the query while radar was not incorporated as the key or value. Our findings indicate that when radar served as the primary information source, supplemented by the fine-grained 3D features of LiDAR, TPCNet achieves superior performance in the prompts containing motion and velocity information. However, for samples with depth-related prompts, the fusion method that uses LiDAR as the query outperforms the radar-query configuration. 
Among the three approaches, BACA demonstrates the best performance across all prompt types. This result highlights the effectiveness of our proposed BACA module in leveraging bidirectional fusion to enrich the environmental context features required by each modality, thereby enhancing its ability to accurately identify the object corresponding to the textual prompt.

\begin{figure}
    \centering
    \includegraphics[width=0.998\linewidth]{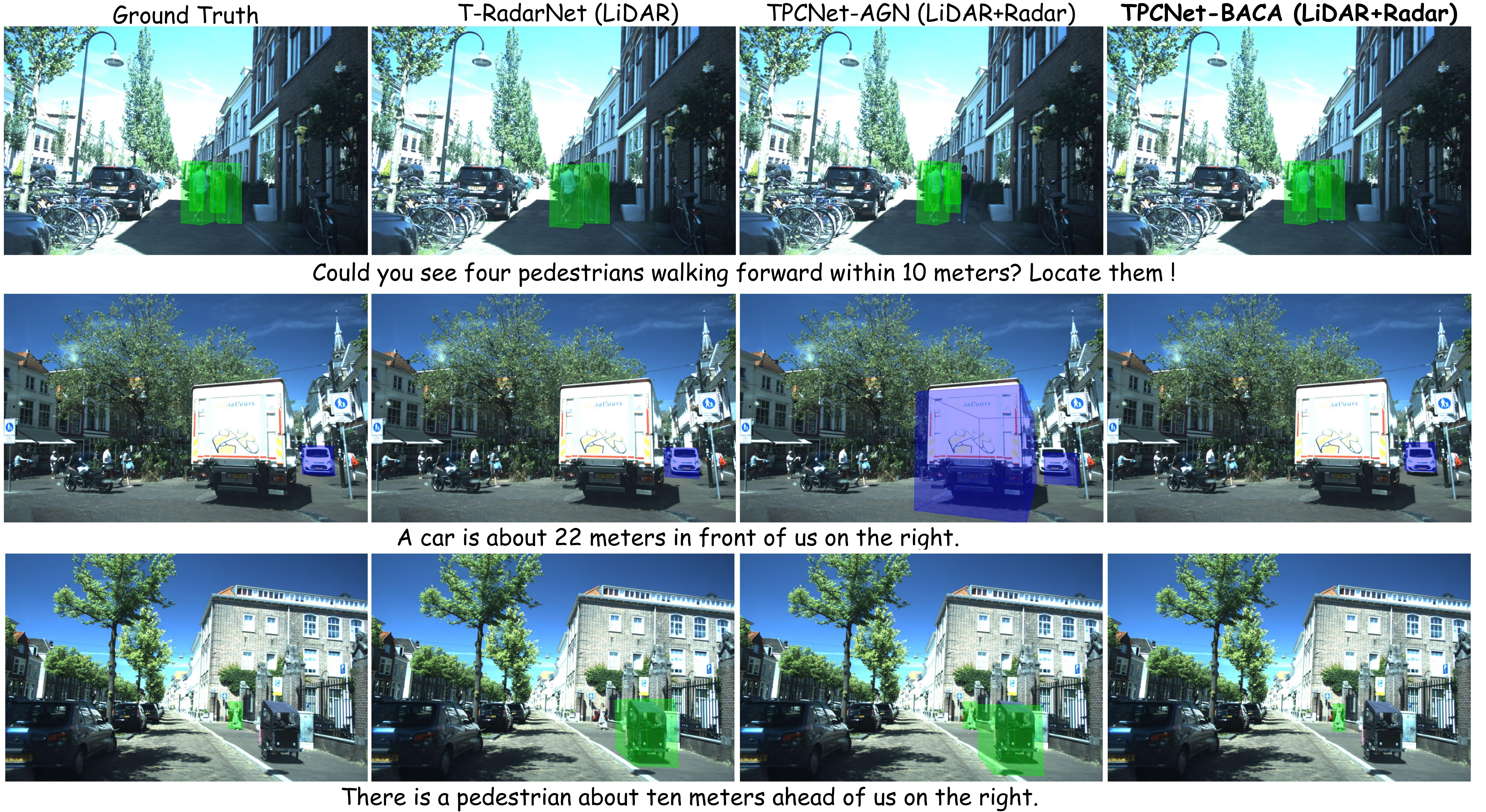}
    \vspace{-6mm}
    \caption{Comparison of various models from the views of RGB camera.}
    \label{fig:visual_compare}
\end{figure}

\begin{figure}
    \centering
    \includegraphics[width=0.99\linewidth]{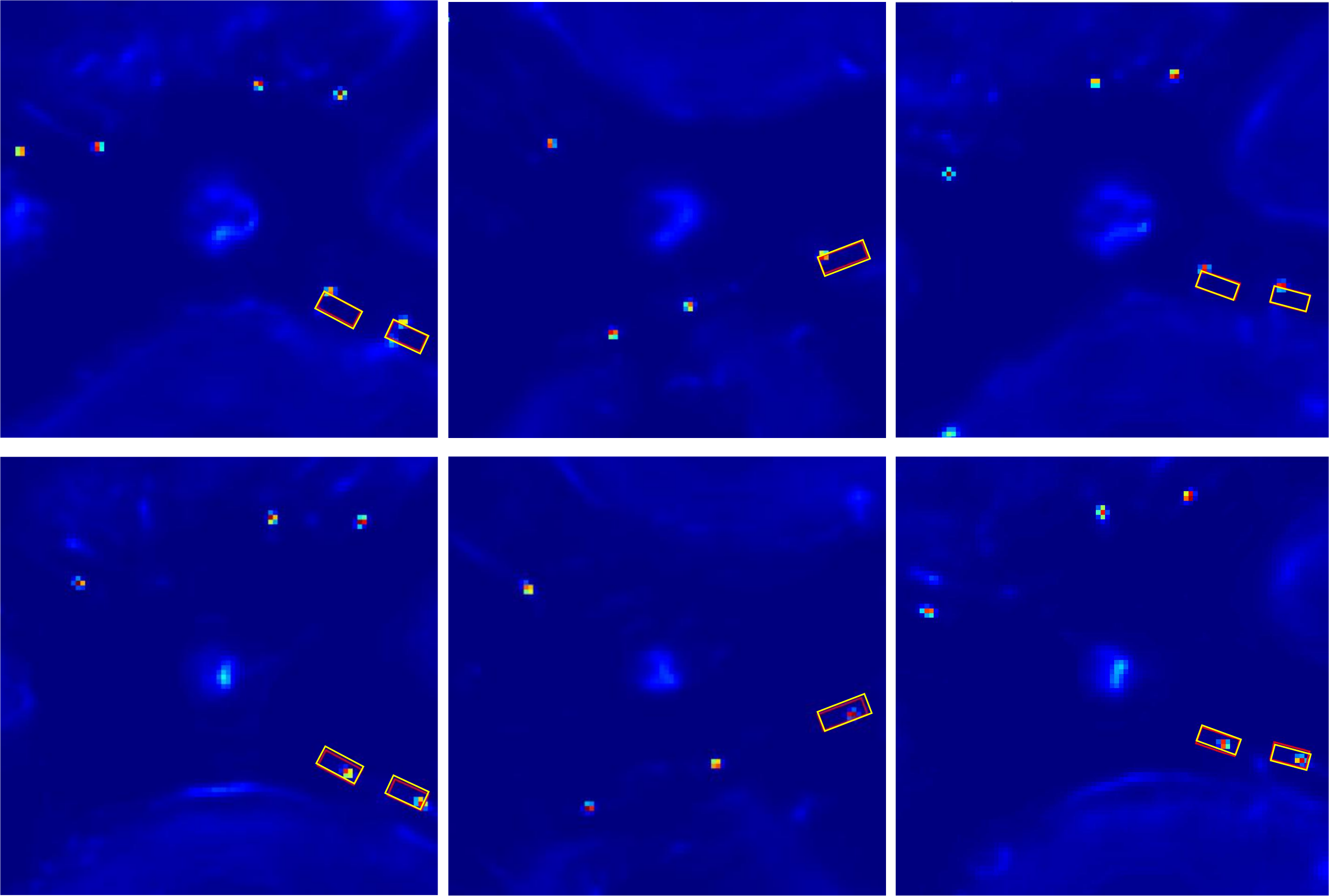}
    \vspace{-4mm}
    \caption{Heatmaps of our proposed C3D-RECHead (first row) and CenterPoint head (second row). Red bounding boxes denote the predicted objects while yellow bounding boxes denote ground truth.}
    \label{fig:prediction_heatmap}
\end{figure}

\subsection{Visualization and Discussion}
Fig. \ref{fig:overall_visualization} illustrates the prediction of TPCNet on the Talk2Radar dataset. As shown, TPCNet effectively captures the objects of varying sizes corresponding to textual prompts in different scenarios, even in complex environments with clutter. TPCNet demonstrates robust performance in detecting both single and multiple objects, as well as objects at varying distances, including those in close proximity and farther away.

Fig. \ref{fig:lidar_radar_compare} shows the performance of TPCNet under three settings: LiDAR-only, radar-only, and fusion of both sensors. For the first sample, the radar-based prediction produces false positive detections, while the LiDAR-based model yields accurate predictions but lacks the level of refinement achieved by sensor fusion. In the second-row sample, our proposed fusion method consistently outperforms the two single-sensor approaches. For the final sample, both the LiDAR-only and radar-only models exhibit varying degrees of false positives due to clutter, whereas the fusion-based approach effectively suppresses false positives and accurately localizes the objects.

Fig. \ref{fig:visual_compare} compares the performance of three models: T-RadarNet, which relies solely on LiDAR input; TPCNet-AGN, which integrates LiDAR and radar features using the AGN module; and TPCNet-BACA, which employs BACA for dual-sensor feature fusion. In the first sample, where the textual prompt refers to four objects, T-RadarNet is only able to detect the two closest objects while failing to localize the two pedestrians further ahead. TPCNet-AGN successfully identifies all four objects but fails to accurately localize one of them. In contrast, our proposed TPCNet-BACA correctly detects and localizes all four prompted objects. For the second sample, where the prompt contains both distance and orientation information, T-RadarNet accurately localizes the correct object. However, TPCNet-AGN mistakenly identifies the truck in front as the object and fails to correctly localize the car. In the third sample, T-RadarNet incorrectly identifies a car on the front right as the referenced object, while TPCNet-AGN produces an additional false positive detection. Overall, our proposed TPCNet-BACA demonstrates superior accuracy in object identification and localization.

As Fig. \ref{fig:prediction_heatmap} shows, we visualize the heatmap of two prediction heads, including the CenterPoint-based prediction head and our proposed C3D-RECHead. We find that our proposed C3D-RECHead successfully focuses on the corner point closest to the ego vehicle, which leads to more precise bounding box predictions compared to the CenterPoint head. The predicted bounding boxes from C3D-RECHead align more closely with the ground truth, demonstrating that performing 3D visual grounding based on the nearest edge yields more stable results than anchoring to the object's center.

\section{Conclusions, Limitations and Future Works}
\label{sec:conclusion}
This paper presents TPCNet, a novel model for 3D visual grounding in autonomous driving and embodied perception, leveraging textual prompts to guide dual-sensor fusion of LiDAR and radar. The framework integrates three key modules: (i) the Bidirectional Agent Cross Attention (BACA) module for efficient feature fusion at low computational cost, (ii) the Dynamic Gated Graph Fusion (DGGF) module for adaptive graph-based feature selection, and (iii) the C3D-RECHead module for enhanced localization accuracy using depth-aligned prompts. Collectively, these designs enable TPCNet to achieve state-of-the-art performance on both the Talk2Radar and Talk2Car datasets, surpassing existing approaches.

Despite these advances, several limitations remain. Current 3D visual grounding methods, including TPCNet, cannot fully capture contextual scene information or color attributes due to the absence of RGB image features. This constraint reduces the ability to reason about appearance and fine-grained semantic cues in complex driving environments.

Although dual-sensor systems combining LiDAR and radar can address the majority of challenges in 3D visual grounding, certain queries involving object color still require the integration of a camera, even though this inevitably increases system complexity. In the future, we plan to explore a holographic 3D visual grounding framework composed of three sensors. Such integration is expected to provide more flexible and fine-grained object queries, thereby enhancing robustness and adaptability in autonomous driving scenarios.
\ifCLASSOPTIONcaptionsoff
  \newpage
\fi



%
\normalem
\footnotesize
\bibliographystyle{IEEEtran}
\bibliography{bare_jrnl}

%




\end{document}